\documentclass[review]{elsarticle}
\usepackage{lineno,hyperref}
\usepackage{color}
\usepackage{ulem}
\usepackage{comment}
\usepackage{adjustbox}
\usepackage[T1]{fontenc}
\usepackage[utf8]{inputenc}
\usepackage{lineno, hyperref}
\usepackage{graphicx, float}
\usepackage{threeparttable} 
\usepackage{booktabs} 
\usepackage{enumitem}
\usepackage{amsmath,amssymb,amsfonts}
\usepackage[caption = false]{subfig}
\usepackage[table,xcdraw]{xcolor}
\usepackage{tabularx}
\usepackage{arydshln,multirow}
\usepackage{float}
\usepackage{geometry}
\usepackage{tgadventor}
\usepackage[section]{placeins}

\journal{Computers and Electronics in Agriculture}

\biboptions{authoryear}

\begin{document}

\begin{frontmatter}
  \title{Vision Foundation Models in Agriculture: Toward Domain-Specific
    Adaptation for Weed Herbicide Trials Assessment}

  \author[Tecnalia_address,ehu_address]{Leire Benito-Del-Valle}
  \corref{mycorrespondingauthor}\ead{leire.benitodelvalle@tecnalia.com}
  \cortext[mycorrespondingauthor]{Corresponding author at: TECNALIA, Basque
    Research and Technology Alliance (BRTA), Parque Tecnológico de Bizkaia, C/
    Geldo. Edificio 700, E-48160 Derio- Bizkaia, Spain}

  \author[Tecnalia_address,ehu_address]{Artzai Pic\'{o}n}
  \author[Tecnalia_address]{Daniel Mugica}
  \author[BASF_M_address,ehu_address]{Manuel Ramos}
  \author[ehu_address]{Eva Portillo}
  \author[BASF_SP_address]{Javier Romero}
  \author[BASF_SP_address]{Carlos Javier Jimenez}
  \author[BASF_address]{Ram\'{o}n Navarra-Mestre}

  \address[Tecnalia_address]{TECNALIA, Basque Research and Technology Alliance
  (BRTA), Parque Tecnol\'{o}gico de Bizkaia, C/ Geldo. Edificio 700, E-48160
  Derio - Bizkaia (Spain)}
  \address[ehu_address]{University of the Basque Country, Plaza Torres Quevedo,
    48013 Bilbao (Spain)}
  \address[BASF_M_address]{BASF Digital Solutions S.L, Camino Fuente de la
    Mora 1, 28050 Madrid (Spain)}
  \address[BASF_SP_address]{BASF Espanola S.L. Carretera A376, 41710 Utrera,
    Sevilla (Spain)}
  \address[BASF_address]{BASF SE, Speyererstrasse 2, 67117 Limburgerhof
    (Germany)}

  \begin{abstract}

    Herbicide field trials require accurate identification of plant species and
    assessment of herbicide-induced damage across diverse environments. While
    general-purpose vision foundation models have shown promising results in
    complex visual domains, their performance can be limited in agriculture,
    where
    fine-grained distinctions between species and damage types are critical.

    In this work, we adapt a general-purpose vision foundation model to
    herbicide
    trial characterization.
    Trained using a self-supervised learning approach on a large, curated
    agricultural dataset, the model learns rich and transferable
    representations
    optimized for herbicide trials images.

    Our domain-specific model significantly outperforms the best
    general-purpose
    foundation
    model in both species identification (F1 score improvement from 0.91 to
    0.94)
    and damage classification (from 0.26 to 0.33). Under unseen conditions (new
    locations and other time), it achieves even greater gains (species
    identification from 0.56 to 0.66; damage classification from 0.17 to 0.27).
    In
    domain-shift scenarios, such as drone imagery, it maintains strong
    performance
    (species classification from 0.49 to 0.60).

    Additionally, we show that domain-specific pretraining enhances
    segmentation accuracy, particularly in low-annotation regimes. An
    annotation-efficiency analysis reveals that, under unseen conditions, the
    domain-specific
    model achieves 5.4\% higher F1 score than the general-purpose model, while
    using
    80\% fewer labeled samples.

    These results demonstrate the generalization capabilities of
    domain-specific foundation models and their potential to significantly
    reduce manual
    annotation efforts, offering a scalable and automated solution for
    herbicide trial analysis.

  \end{abstract}

  \begin{keyword}
    \begin{footnotesize}
      Foundation Models \sep Large Visual Models \sep Plant Species and Damage
      segmentation \sep Precision Agriculture \sep Precise Phenotyping
    \end{footnotesize}
  \end{keyword}
\end{frontmatter}




\section{Introduction}
\label{sec:introduction}


Weed management is a critical challenge in modern agriculture, directly
impacting crop yields, food security, and economic sustainability
(\cite{kumar_weed_2024}). Weeds compete with crops for essential resources like
soil, nutrients and sunlight, ultimately slowing growth and reducing yields.
Moreover, many weed species serve as hosts for pests and pathogens, further
compromising crop health.

To manage weed populations and optimize crop yield, traditional weed management
methods are commonly employed, such as manual removal or broad-spectrum
herbicide application. These methods, however, are labor-intensive, costly and
environmentally unsustainable (\cite{oerke_crop_2006}). Thus, accurate weed
detection and segmentation in crops are essential for precision agriculture,
enabling targeted interventions that minimize herbicide use and reduce
environmental impact. Automated weed detection systems, utilizing technologies
such as deep learning and computer vision ease the identification and
management of weeds, reducing reliance on manual labor (\cite{hu_deep_2024}).
Furthermore, precision spraying technologies can reduce herbicide usage,
thereby diminishing environmental contamination
(\cite{darbyshire_towards_2023}).

Initially, machine learning (ML) approaches have been employed to classify or
detect crops and weeds, relying on manual feature engineering techniques such
as texture and color histograms (\cite{juwono_machine_2023}). These methods
often lack generalizability across varying field conditions and crop types due
to their dependence on handcrafted features
(\cite{adhinata_comprehensive_2024}). Therefore, the introduction of deep
learning (DL) has marked a turning point, enabling end-to-end learning from raw
imagery and significantly improving performance in all tasks.

\cite{mortensen_semantic_2016} first applied deep learning to crop and weed
semantic segmentation, achieving 79\% pixel accuracy for plant species. They
later improved the method, enabling the discernment of corn from 23 weed
species with a 94\% pixel-wise accuracy (\cite{dyrmann_pixel-wise_2016}).

However, like many DL-based methods, these approaches heavily rely on large,
manually annotated datasets. When developing effective weed detection systems,
large, diverse, and accurately annotated datasets are needed to train precise
deep learning algorithms. Acquiring such datasets, however, is time-consuming,
costly and often impractical. For instance, the progression of diseases and
weed growth in natural environments is prone to significant changes, making it
hard to capture a large variability. Consequently, these models tend to perform
poorly in the underrepresented regions or require mayor redesign to incorporate
novel weed species.

To reduce the burden of manual labeling, various strategies have been proposed.
\cite{bosilj_transfer_2020} have shown that transfer learning can significantly
reduce annotation demands while maintaining high accuracy in crop and weed
discrimination. Other approaches have focused on architectural improvements,
such as weedNet (\cite{sa_weednet_2018}), which achieves an 80\% F1 score using
an encoder-decoder cascaded deep neural network (Segnet), and
\cite{milioto_real-time_2018}, who integrate task-relevant background knowledge
into the network to improve generalization and real-time inference performance.
Similarly, \cite{ma_fully_2019} employ a fully convolutional network for rice
and weed segmentation, reporting an average accuracy of 92.7\%.

Another setback when developing effective weed detection systems is
distinguishing weeds from crops. This is primarily due
to the fine-grained visual similarities between weeds and crops, especially
during early growth stages, as well as the significant intra-class variability
among weed species. To overcome this,  probabilistic methods have been
suggested for uncertainty estimation (\cite{celikkan_semantic_2023}).

These prior approaches have targeted the identification and segmentation of
healthy plants. While this is effective for precision agriculture applications,
it does not fully address the needs of automating herbicide research and
development. In these field trials, it is necessary to not only distinguish
between healthy plant species, but also to segment damaged plants and quantify
different types of damage for each species. This adds further complexity to the
problem.

Given the fine-grained nature of plant species identification, recent research
has explored how incorporating taxonomic relationships among species can
enhance model performance. \cite{picon_taxonomic_2025} propose a
hierarchical loss function for semantic segmentation tasks that leverages
taxonomic relationships between species, plant damages, and other related
factors, allowing models to benefit from partially annotated data.

While this approach has demonstrated strong performance in field
trials, effectively addressing key challenges in crop-weed discrimination and
damage quantification, its evaluation has been limited to in-domain data. As a
result, its ability to generalize to unseen conditions remains unclear. This
limitation restricts its applicability in real-world scenarios, where
environmental variability and species diversity are prevalent.

Recently, foundation models have emerged, which address the growing demand for
large-scale datasets. These models are trained on vast amounts of data and are
capable of producing representations that can be applied across a wide range of
tasks and domains (\cite{bommasani_opportunities_2022}). A particularly
promising approach involves training foundation models using self-supervised
learning (SSL), which is especially useful in areas like agriculture where
publicly available image data is scarce. SSL works by altering data in a way
that creates implicit labels, allowing models to learn without explicit
annotations.

A notable example is DINOv2 (\cite{oquab_dinov2_2024}), a vision foundation
model trained using SSL techniques that learns robust and transferable visual
features without supervision. DINOv2 has shown strong performance across
various downstream tasks, and has shown great potential in agriculture.
\cite{picon_field_2025} applied DINOv2 for weed and damage segmentation in
herbicide trials, demonstrating its robustness under domain shifts. Similarly,
\cite{espejo-garcia_foundation_2025} evaluated DINOv2's adaptability across
multiple agricultural datasets, highlighting its effectiveness in disease
detection, weed identification, and growth stage classification, even with
minimal fine-tuning.

Building on this, DINOv3 (\cite{simeoni_dinov3_2025}) introduces further
improvements by scaling both the data and the model size, and by incorporating
novel techniques such as Gram anchoring. It (DINOv3\_web) achieves
state-of-the-art results across diverse vision tasks without fine-tuning and
introduces variants tailored for satellite imagery (DINOv3\_sat), showcasing
the potential of domain-specific adaptations.

Despite these advancements, most large vision foundation models remain
general-purpose and are not adapted to specialized domains like agriculture.
In particular, herbicide trials present unique challenges, such as high
intra-species variability and the need for fine-grained damage assessment, that
continue to pose difficulties for existing models. Building on these
observations, we propose that domain-specific adaptations, particularly those
leveraging self-supervised learning, could lead to significant performance
improvements in real-world agricultural settings.

The goal of this work is to develop a domain-specific large vision foundation
model for herbicide trials, using a self-supervised learning approach. By
tailoring the model to agricultural imagery, we aim to reduce the need for
annotated data and improve generalization across diverse real-world conditions.

The main contributions are as follows:
\begin{itemize}
  \item A domain-specific vision foundation model for herbicide trials, is
        proposed. It is trained using a self-supervised learning approach on a
        large curated dataset, enabling the model to learn rich and
        transferable representations tailored to agricultural imagery.
  \item A comparative evaluation is performed against state-of-the-art
        general-purpose foundation models on multi-species and damage plant
        semantic segmentation tasks. This comparison highlights the benefits of
        domain-specific pretraining for agricultural applications.
  \item The robustness of the model is evaluated under domain shifts,
        including changes in camera types, geographic locations, and capture
        protocols, confirming its generalization capabilities in real-world
        agricultural scenarios.
  \item A detailed annotation-efficiency analysis is performed. By
        gradually decreasing the amount of labeled training data, we assess
        the model's performance in low-annotation regimes and its potential to
        reduce reliance on manual labeling.

\end{itemize}

\section{Materials}
\label{sec:materials}

In this section, we describe the datasets employed for the generation and
evaluation of the domain-specific foundation model. Each dataset serves a
distinct purpose in the training and evaluation pipeline (see
Section~\ref{sec:methods} for a detailed explanation of the pipeline):
\begin{itemize}
  \item Herbicide trials dataset: used to train the domain-specific foundation
        model through self-supervised learning.
  \item Primitives dataset: employed to monitor learning progress during
        self-supervised pretraining.
  \item Base dataset: used for supervised fine-tuning of segmentation decoders
        and performance evaluation.
  \item Reality check dataset: applied to assess model generalization
        capabilities under temporal and environmental domain shifts.
  \item Reality check DRONE dataset: used to evaluate model robustness under
        extreme domain shifts involving different sensor modalities and capture
        protocols.
\end{itemize}

\subsection{Herbicides trials dataset}
\label{subsec:herbicides_trials_dataset}

To develop a domain-specific foundation model, it is essential to gather a
large and well-curated dataset. For this purpose, we compiled all available
herbicide trial images (BASF Agricultural Solutions), resulting in an initial
dataset
comprising 340,554 images. These were acquired using diverse sensors across
multiple geographic locations, years, and crop types, ensuring substantial
diversity in environmental conditions and visual characteristics.

Although the dataset offers broad variability, its quality cannot be guaranteed
without further inspection. Therefore, an initial data curation step was
performed to identify and remove low-quality samples. Specifically, blurry
images, exact duplicates, near-duplicates, and excessively dark samples were
discarded, reducing the dataset to 261,891 images.

From these curated images, we generated non-overlapping tiles of 518x518
pixels, yielding a total of 11.8M image patches. The large number of tiles is
attributed to the high resolution and extensive field coverage of
the original images, which often contain multiple crop rows and treatment
zones.

\begin{figure*}[ht]
  \centering
  \includegraphics[width=15cm]{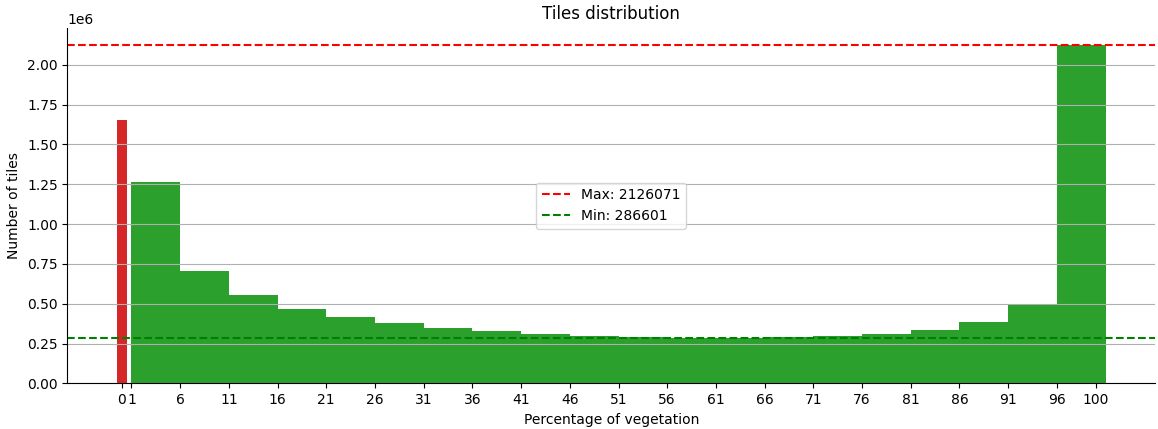}
  \caption{Distribution of image tiles based on vegetation coverage.}
  \label{fig:tiles_distribution}
\end{figure*}

Following tiling, we observed substantial variation in vegetation coverage
across the dataset. To ensure that the model learns meaningful representations
of vegetation presence and structure, rather than relying on superficial cues
such as dominant color or background texture, we performed a balancing step
based on vegetation content. Using segmentation masks generated by an existing
vegetation segmentation model, we computed the percentage of vegetation present
in each tile. The distribution of vegetation coverage is shown in Figure
\ref{fig:tiles_distribution}, revealing a strong imbalance.

To address this, we used the predefined vegetation coverage intervals shown in
Figure \ref{fig:tiles_distribution} and selected an equal number of tiles from
each interval. This approach ensures a uniform representation of vegetation
density across the dataset, promoting more robust and generalizable feature
learning during model training.

\begin{figure*}[ht]
  \centering
  \includegraphics[width=15cm]{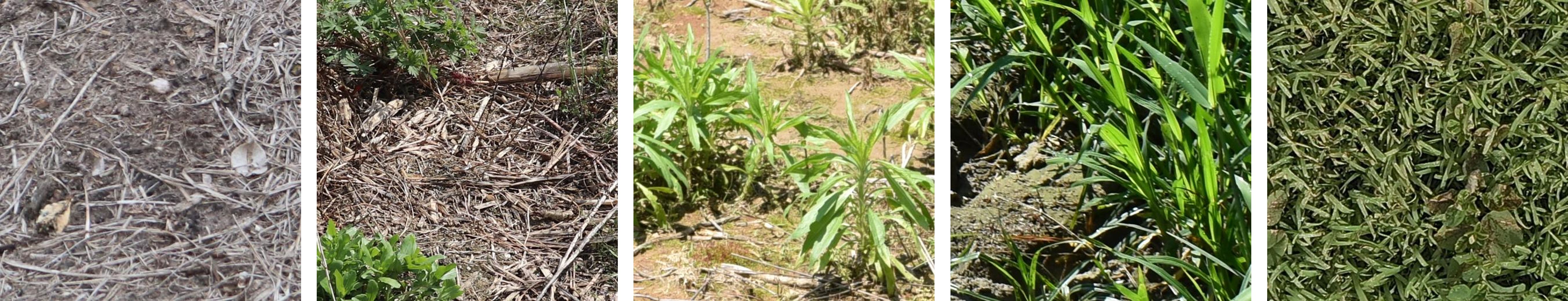}
  \caption{Image samples from the curated herbicides trials dataset.}
  \label{fig:examples_tiles_dataset}
\end{figure*}

After this final balancing step, we obtained a curated dataset of 6M image
tiles (see Figure \ref{fig:examples_tiles_dataset}), which was used to
train the domain-specific foundation model.

\subsection{Base dataset}
\label{subsec:base_data}

For the Base dataset, we employed a reduced version of the dataset presented in
\cite{picon_taxonomic_2025}, where a carefully compiled and manually annotated
dataset was generated between 2018 and 2020. This dataset comprises a wide
range of crop and weed species collected from real  plot fields over multiple
years. The 14 species included are:
\textit{Abutilon theophrasti} (eppo: ABUTH),
\textit{Amaranthus retroflexus} (eppo: AMARE),
\textit{Chenopodium album} (eppo: CHEAL),
\textit{Digitaria sanguinalis} (eppo: DIGSA),
\textit{Echinochloa crus-galli} (eppo: ECHCG),
\textit{Portulaca oleracea} (eppo: POROL),
\textit{Setaria viridis} (eppo: SETVE),
\textit{Datura stramonium} (eppo: DATST),
\textit{Echinochloa colona} (eppo: ECHCO),
\textit{Glycine max} (eppo: GLXMA),
\textit{Helianthus annuus} (eppo: HELAN),
\textit{Polygonum convolvulus} (eppo: POLCO),
\textit{Solanum nigrum} (eppo: SOLNI),
and maize, \textit{Zea mays} (eppo: ZEAMX).

The dataset features images depicting 6 types of plant damage, such as
INITIAL, INITIAL-BLEACHING, INITIAL-NECROSIS, BLEACHING, NECROSIS, and
LEAF-CURLING, along with annotated growth stages. These images were captured
under diverse environmental conditions and locations. Moreover, two dataset
formats were developed: type A, which contains real field images with
multiple species, and type C, in which only individual species are present.

In this study, we focus exclusively on Type A datasets from 2019, representing
a reduction of 82.48\% in data size (from 2449 to 429). This choice aligns with
our objective of minimizing the amount of labeled data required for effective
model training.

Annotation was performed using CVAT (\cite{cvat}), and further refined using
existing vegetation segmentation models. Metadata such as location and image
field of view were included to support scale correction and data augmentation.
All species are labeled using their respective EPPO codes
(\cite{ayllon2023eppo}).

Species abundance and distribution vary across plots due to environmental
factors and species-specific traits, resulting in a rich diversity within the
dataset. A detailed overview of the contents of the dataset is provided in
Table \ref{tab:real_field_datasets}.

\begin{table}[ht]
  \centering
  \resizebox{\textwidth}{!}{%

    \begin{tabular}{p{0.08\linewidth}p{0.04\linewidth}p{0.06\linewidth}p{0.34\linewidth}p{0.20\linewidth}p{0.15\linewidth}}
      \hline
      Collection                               & Type
                                               & Images                      &
      Species                                  &
      Damage                                   & Location
      \\
      \hline
      2019A1                                   & A
                                               & 147                         &
      ABUTH, AMARE, CHEAL, DATST, DIGSA, ECHCG, ECHCO, GLXMA,
      HELAN, POLCO, POROL, SETVE, SOLNI, ZEAMX & INITIAL, INITIAL-BLEACHING,
      INITIAL-NECROSIS, BLEACHING, NECROSIS    & ES, DE

      \\
      2019A2                                   & A
                                               & 282                         &
      ABUTH, AMARE, CHEAL, DATST, DIGSA, ECHCG, ECHCO, GLXMA,
      HELAN, POLCO, POROL, SETVE, SOLNI, ZEAMX & INITIAL-BLEACHING,
      INITIAL-NECROSIS,
      BLEACHING, NECROSIS                      & ES, DE

      \\
      \hline
    \end{tabular}
  }
  \caption{BASE dataset content summary.}
  \label{tab:real_field_datasets}
\end{table}

\begin{figure*}[ht]
  \centering
  \includegraphics[width=14cm]{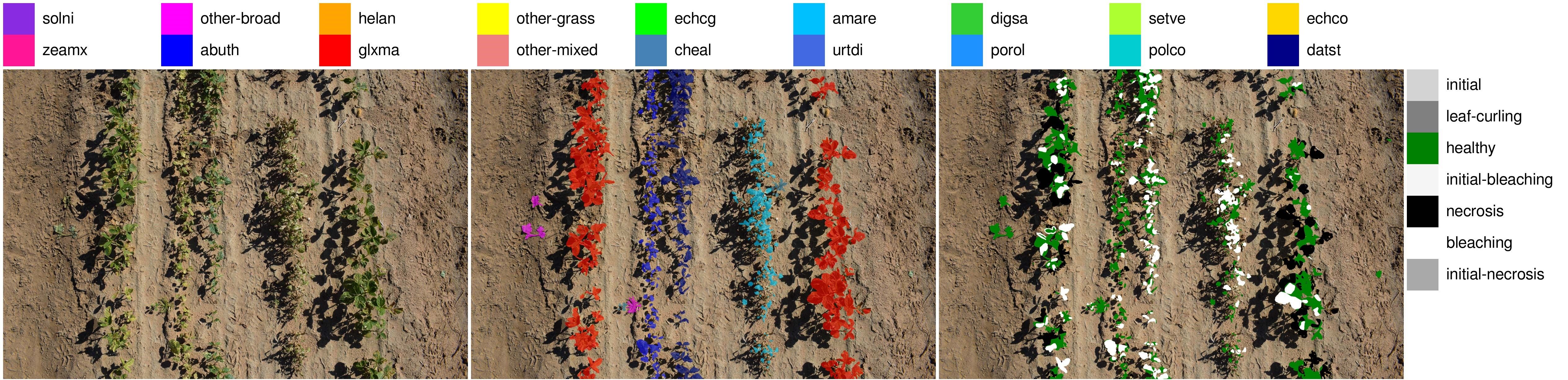}
  \caption{2019A2 dataset. Example image with species, and damage masks. left)
    Original Image, middle) Species annotation, right) Damage annotation
    (\cite{picon_taxonomic_2025}).}
  \label{fig:2019A2_fig4_1}
\end{figure*}

An example of the annotated images is shown in Figure \ref{fig:2019A2_fig4_1},
which illustrates how species, and damage types are labeled.
Given the inherent uncertainty in the annotation process, we introduced
additional class labels to handle ambiguous cases. These include categories for
unidentified broadleaf species (other-broad), unidentified grass species
(other-grass), and instances where multiple species are present (other-mixed).

\subsection{Primitives dataset}
\label{subsec:primitives_dataset}

To evaluate the performance of the foundation model during training (SSL) and
to select the optimal weights, we generated an auxiliary species classification
dataset. This dataset is composed of primitives extracted from the original
Base dataset described in \cite{picon_taxonomic_2025}. For this purpose, we
exclusively used the C-type subsets, as they contain images with a single
species per image, ensuring consistency in labeling (Section
\ref{subsec:base_data}). A detailed overview of the
full images used for the primitive extraction is provided in Table
\ref{tab:real_field_datasets_primitives}.

\begin{figure*}[ht]
  \centering
  \includegraphics[width=13cm]{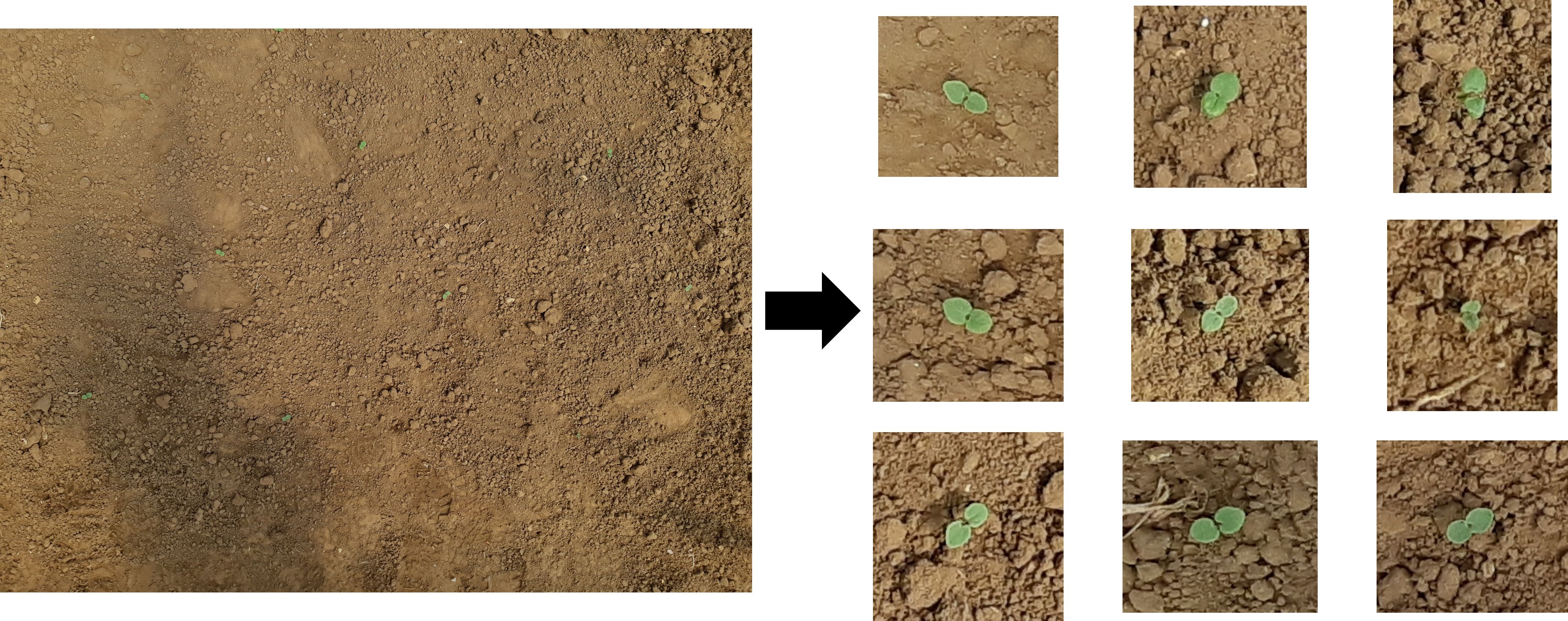}
  \caption{Primitive extraction example showing the original image (left) and
    the corresponding cropped regions (right) used as individual primitives.}
  \label{fig:primitives}
\end{figure*}

The primitives dataset was constructed using the species semantic segmentation
masks available at the image level. Each segmented blob was isolated to produce
individual images representing distinct plant instances. Due to the simplicity
of the cropping process, some primitives may contain overlapping plants.
However, since all plants within a given image belong to the same species, this
does not compromise the suitability of the dataset for classification tasks. An
example of the primitives extracted from a single image is shown in Figure
\ref{fig:primitives}.

\begin{table}[ht]
  \centering
  \resizebox{\textwidth}{!}{%

    \begin{tabular}{p{0.08\linewidth}p{0.04\linewidth}p{0.06\linewidth}p{0.34\linewidth}p{0.20\linewidth}p{0.15\linewidth}}
      \hline
      Collection                               & Type
                                               & Images               & Species
                                               & Damage
                                               & Location
      \\
      \hline
      2019C1                                   & C
                                               & 249                  & ABUTH,
      AMARE, CHEAL, DATST, DIGSA, ECHCG, GLXMA, HELAN,
      POLCO, POROL, SETVE, SOLNI, ZEAMX        & None
                                               & ES
      \\
      2019C2                                   & C
                                               & 303                  & ABUTH,
      AMARE, CHEAL, DATST, DIGSA, ECHCG, GLXMA, HELAN,
      POLCO, POROL, SETVE, SOLNI, ZEAMX        & NECROSIS
                                               & ES
      \\
      2020CDE1                                 & C
                                               & 139                  & ABUTH,
      AMARE, CHEAL, DIGSA, ECHCG, POLCO, POROL &
      NECROSIS                                 & DE

      \\
      2020CES1                                 & C
                                               & 222                  & ABUTH,
      AMARE, CHEAL, DATST, DIGSA, ECHCG, ECHCO,
      GLXMA, HELAN, POLCO, POROL, SETVE, SOLNI & BLEACHING, NECROSIS,
      LEAF-CURLING                             &
      ES

      \\
      \hline
    \end{tabular}
  }
  \caption{BASE dataset content used to generate the primitives dataset.}
  \label{tab:real_field_datasets_primitives}
\end{table}

\subsection{Reality check (2023)}
\label{subsec:reality_check_dataset}

To assess the algorithm's performance in unfamiliar scenarios, we utilized the
same reality check dataset introduced in \cite{picon_field_2025}. This dataset
was collected in 2023 across Germany, the United States, and Spain. The images
were acquired using mobile and digital cameras through the BASF crop
digitalization app and annotated using CVAT (\cite{cvat}). It includes 35
identified crop and weed species, as well as broader categories for
unidentified vegetation and species not present in the Base dataset
(Section \ref{subsec:base_data}). These unclassified or novel species are
grouped
according to their botanical taxonomy into three categories: other-broad,
other-grass, and other-mixed. Moreover, some species may not be present in this
dataset due to environmental changes or other circunstances. A detailed
overview
of the contents of the dataset is provided in Table \ref{tab:reality_dataset}.

\begin{table}[ht]
  \centering
  \resizebox{\textwidth}{!}{%

    \begin{tabular}{p{0.08\linewidth}p{0.04\linewidth}p{0.06\linewidth}p{0.34\linewidth}p{0.20\linewidth}p{0.15\linewidth}}
      \hline
      Collection                                       & Type
                                                       & Images              &
      Species
                                                       & Damage              &
      Location
      \\
      \hline
      2023ADE                                          & A
                                                       & 32                  &
      AMACH, CHEAL, DATST, ECHCG, GLXMA, HELAN,
      HORVX, LAMAM,
      LAMPU, LOLMU, MERAN, POLAM, POLCO, POLPE, POROL, SETVE, SETVI, SINAL,
      SPRAR,
      STEME, TRZAX, ZEAMX,                             & INITIAL, BLEACHING,
      NECROSIS, LEAF-CURLING                           & GERMANY
      \\

      2023AES                                          & A
                                                       & 36                  &
      AMABL, AMARE,  CHEAL, DATST, DIGSA,
      ECHCG, GLXMA,
      HELAN, POLCO, POROL, SETVE, SETVI, SOLNI, ZEAMX, & INITIAL, NECROSIS,
      LEAF-CURLING                                     & SPAIN

      \\

      2023AUS                                          & A
                                                       & 37                  &
      AMABL, AMAPA,  AMARE, AMATU, CHEAL,
      CONAR, CYPES,
      DIGSA, ECHCG, ELEIN, GLXMA, HELAN, KCHSC, MOLVE, POLPY, POROL, SETPU,
      SETVI,
      ZEAMX,                                           & INITIAL, BLEACHING,
      NECROSIS, LEAF-CURLING                           & UNITED STATES
      \\

      \hline
    \end{tabular}
  }
  \caption{REALITY dataset content summary.}
  \label{tab:reality_dataset}
\end{table}

\subsection{Reality check DRONE based dataset (2024)}
\label{subsec:drones_dataset}

To further evaluate the algorithm under conditions of significant data drift,
we used the drone-based dataset introduced in \cite{picon_field_2025}.	It was
collected between late 2023 and 2024 in Spain, Germany, and the United States
using DJI drones equipped with RGB sensors. The dataset includes images of
maize (\textit{Zea mays}, ZEAMX), sunflower (\textit{Helianthus annuus},
HELAN), and soybean (\textit{Glycine max}, GLXMA). A summary of the
dataset is presented in Table \ref{tab:reality_dataset_drones}.

\begin{table}[ht]
  \centering
  \resizebox{\textwidth}{!}{%

    \begin{tabular}{p{0.08\linewidth}p{0.04\linewidth}p{0.06\linewidth}p{0.34\linewidth}p{0.20\linewidth}p{0.15\linewidth}}
      \hline
      Collection                               & Type
                                               & Images
                                               & Species
                                               & Damage
                                               & Location
      \\
      \hline
      UAV241122                                & A
                                               & 201
                                               & ABUTH, ACCOS, ALOMY, AMABL,
      AMACH, AMAPA, AMARE,
      AMATA, AMATU, ANGAR, AVEFA, BRSNN, CAPBP, CASOB, CHEAL, CHEHY, CIRAR,
      CONAR,
      CYPCP, CYPES, DATST, DIGSA, DTTAE, DTTSS, ECAEL, ECHCG, ECHCO, ELEIN,
      GALAP,
      GLXMA, GOSHI, HELAN, HISIN, HORVX, IPOHE, IPOLA, KCHSC, LAMAM, LAMPU,
      LOLMU,
      MERAN, MOLVE, OTHER-BROAD, OTHER-GRASS, OTHER-MIXED, PANDI, PANMI, PESGL,
      PIBSA, POLAM, POLCO, POLPE, POLPY, POROL, PORPI, RAPRA, RCHSC, SENVU,
      SETPU,
      SETVE, SETVI, SIDSP, SINAL, SOIL, SOLNI, SOLTU, SONOL, SORVU, SPRAR,
      STEME,
      TRBTE, TRZAX, URTDI, URTUR, VIOAR, ZEAMX & INITIAL, BLEACHING, NECROSIS,
      LEAF-CURLING                             & GERMANY, UNITED STATES,
      SPAIN,
      \\
      \hline
    \end{tabular}
  }
  \caption{DRONE based reality dataset content summary.}
  \label{tab:reality_dataset_drones}
\end{table}

\section{ Methods}
\label{sec:methods}

\begin{figure*}[ht]
  \centering
  \includegraphics[width=8.5cm]{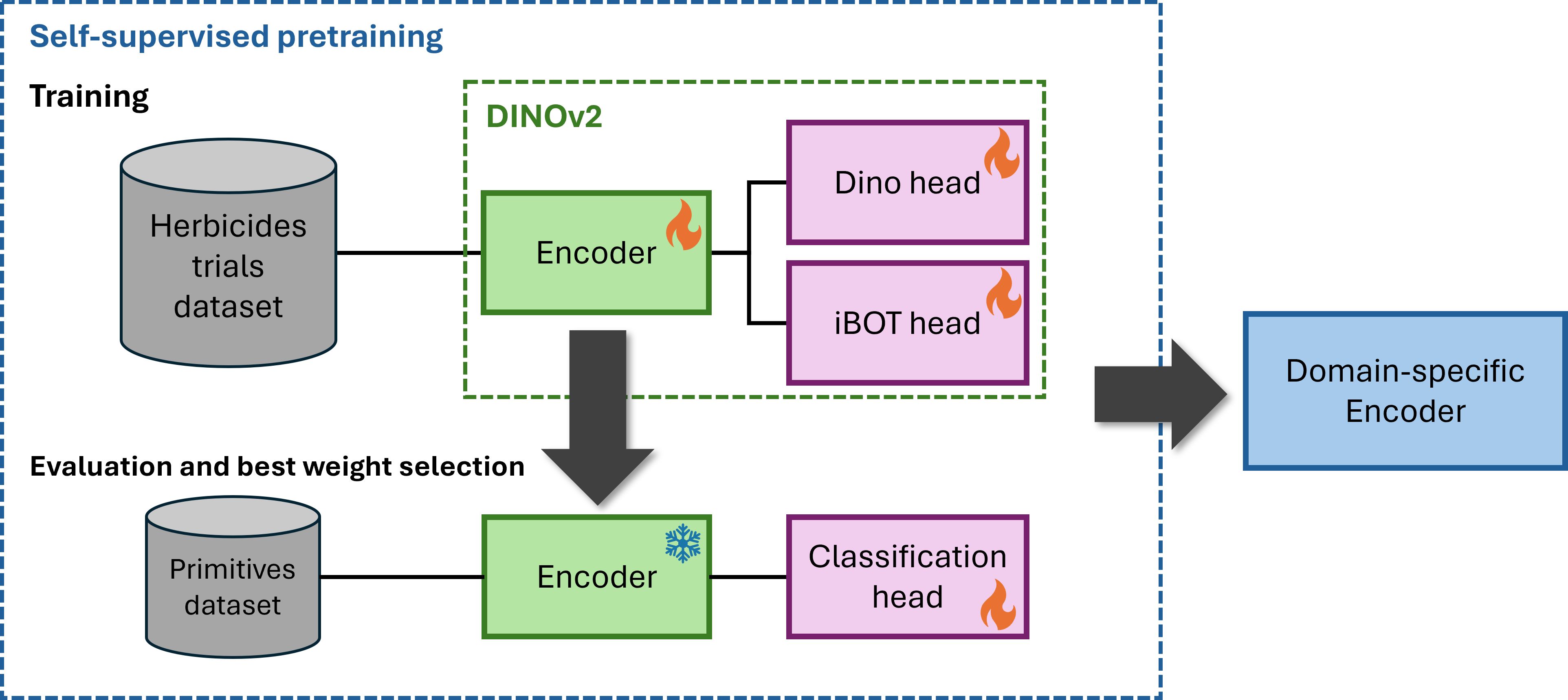}
  \includegraphics[width=8.5cm]{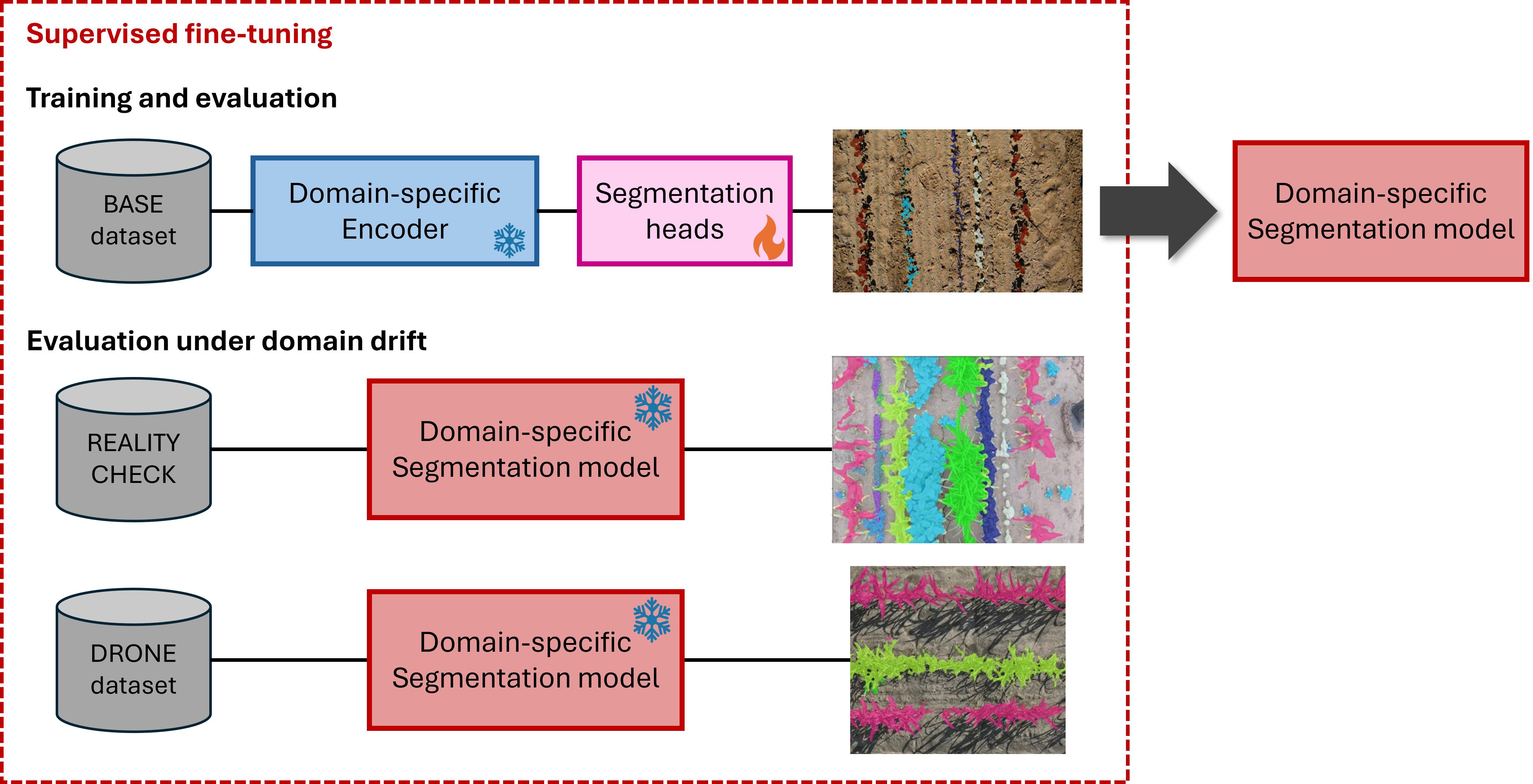}
  \caption{Overview of the proposed methodology. Left) Self-supervised
    pretraining stage where DINOv2 is fine-tuned on the herbicides trials
    dataset
    (Section \ref{subsec:herbicides_trials_dataset}). Model performance is
    monitored and optimal weights are selected using the primitives dataset
    (Section \ref{subsec:primitives_dataset}). Right) Supervised
    fine-tuning stage where the segmentation model is trained on the BASE
    dataset
    (Section \ref{subsec:base_data}). Model evaluation is performed on the
    test subset of the BASE dataset
    (same domain) and two additional datasets to assess domain shift
    robustness: REALITY CHECK (Section
    \ref{subsec:reality_check_dataset})
    and DRONE dataset (Section \ref{subsec:drones_dataset}).}
  \label{fig:pipeline}
\end{figure*}

This section introduces the methodology followed to obtain a domain-specific
foundation model. The process is structured in two main stages (see Figure
\ref{fig:pipeline}): self-supervised
pretraining, in which the foundation model is formed by learning general
domain representations from unlabeled data, and supervised fine-tuning, which
serves to validate the model's effectiveness on a downstream segmentation task.

\subsection{Self-supervised pretraining}

To obtain the domain-specific foundation model, we employed the large variant
of DINOv2 with registers (ViT-L/14, 0.3B parameters)(\cite{oquab_dinov2_2024}),
a self-supervised learning framework based on vision transformers (ViT)
(\cite{dosovitskiy_image_2021}). DINOv2 operates with a dual-network
architecture, where a student network is trained to align with the output of a
teacher network, whose weights are updated via an exponential moving average of
the student. Images are split into patches and processed with a Vision
Transformer backbone, followed by projection heads that produce feature
representations. The training objective encourages the student to match the
teacher's output across augmented views, enabling stable self-supervised
learning without labels.

The model was trained on the herbicides trials dataset (see Section
\ref{subsec:herbicides_trials_dataset}) using the official
DINOv2 implementation, starting from publicly available pretrained weights for
the backbone. Since the DINOv2-specific heads are not released, we initialized
them randomly and followed a two-phase training strategy: first training only
the heads for 20 epochs, followed by full model training for 1000 epochs.

We evaluated two input resolutions during training: 518x518 pixels, which
matches both the default DINOv2 configuration and the resolution of our
dataset, and 224x224 pixels, which allowed for a larger global batch size. For
the latter, the positional encodings were adapted to the resolution using
bicubic interpolation.

Training was performed on 6 NVIDIA Tesla H100 GPUs (80GB each). The global
batch size was set to the maximum each GPU could handle without running
out of memory: 144 for 518x518 inputs and 360 for 224x224 inputs. The 518x518
resolution matches the original DINOv2 implementation, whereas the 224x224
resolution was selected to enable a batch size closer to the smallest batch
size tested in the original implementation (2048) (\cite{oquab_dinov2_2024}).
During head-only training, the base learning rate was set to $2 \times 10^{-3}$
with 5 warmup epochs. For full model training, the learning rate was reduced to
$2 \times 10^{-4}$ with 10 warmup epochs. All other hyperparameters were kept
at their default values.

To monitor learning progress during self-supervised training, model weights
were saved every 5 epochs. These checkpoints were evaluated using a linear
classification protocol on the primitives dataset (see Section
\ref{subsec:primitives_dataset}). The final model weights were
selected based on the epoch that achieved the highest classification accuracy,
ensuring that the retained foundation model reflects the best representation
quality learned during training. Therefore, this stage resulted in two
domain-specific models: DINOv2\_FT\_518 and DINOv2\_FT\_224, corresponding to
the two input resolutions used.

\subsection{Supervised fine-tuning}
\label{subsec:sup_ft}

For supervised fine-tuning, we adopted the same architecture and training
procedure described in \cite{picon_field_2025}. The generated domain-specific
foundation model was used as the encoder backbone (DINOv2) within a
segmentation framework composed of task-specific decoders. The decoder design
follows the SegFormer multi-scale architecture (\cite{xie2021segformer}),
allowing efficient integration of features accross different spatial scales.

\begin{figure*}[ht]
  \centering
  \includegraphics[width=12.5cm]{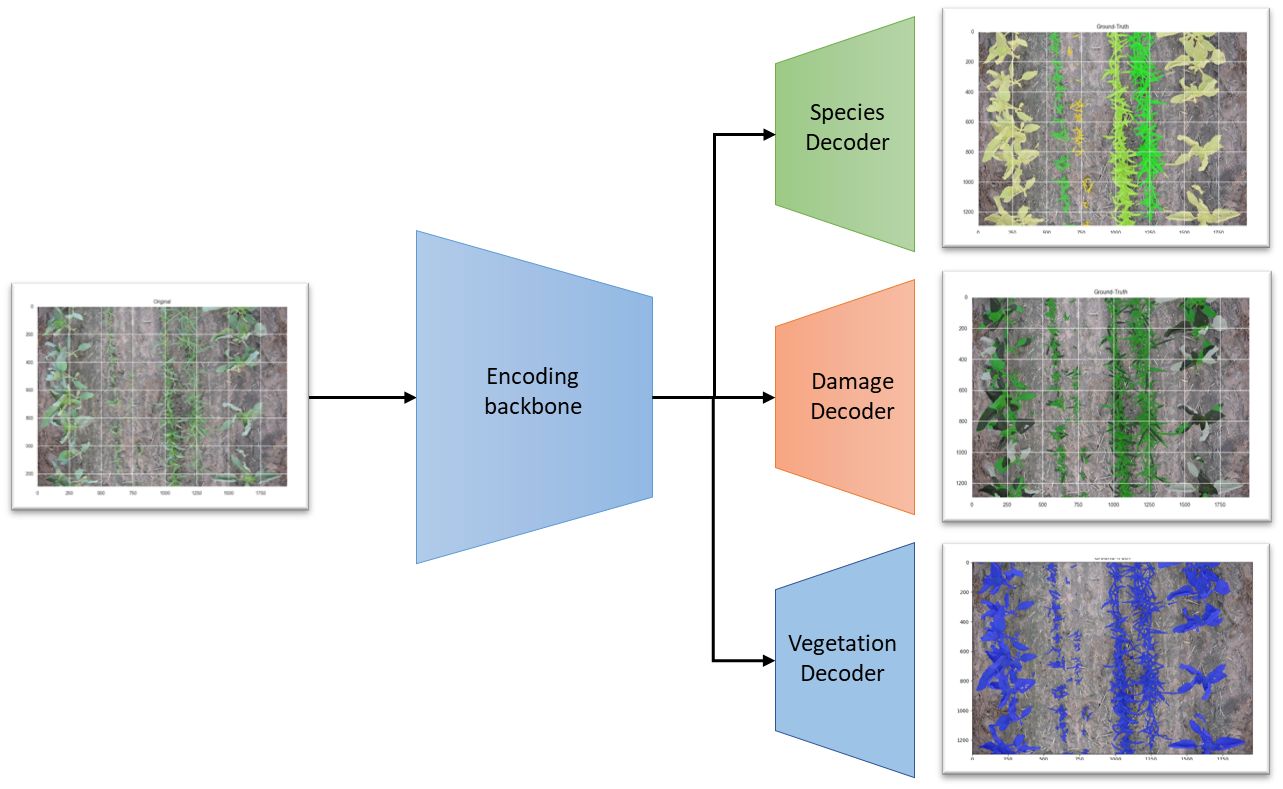}
  \caption{Proposed architecture. An encoding backbone (all the evaluated DINO
    variants) is connected into three independent decoders. Each decoder covers
    one of the network tasks (species, damage, and vegetation)
    (\cite{picon_taxonomic_2025}).}
  \label{fig:diagram_architectures}
\end{figure*}

Three independent decoders were used for the tasks of vegetation presence,
species identification, and damage classification, as illuestrated in Figure
\ref{fig:diagram_architectures}. The final layer of each decoder was replaced
with a 1x1 convolutional layer, with the number of filters corresponding to the
number of classes in the task. Softmax activation was used for multi-class
outputs. The encoder weights were frozen during training to preserve the
representations learned during the self-supervised pretraining.

Loss functions per task were defined as weighted categorical cross-entropy,
with class weights computed using the effective number of samples method
(\cite{cui2019class}), extended to semantic segmentation. A beta parameter of
0.99 was used to address class imbalance.

While we followed the same augmentation strategies and optimizer configuration
as in \cite{picon_field_2025}, the dataset split was significantly different.
Specifically, training was performed using 80\% of the 2019A2 subset and the
full 2019A1 subset (373 images), with validation on 10\% of
2019A2. The test set remained consistent, using 10\% of 2019A2. This change led
to a substantial reduction in the amount of labeled data used for
training, from 2,393 images in the original setup to just 373, representing an
84.4\% decrease.

Training was conducted on an NVIDIA Tesla H100 GPU with 80GB of memory.
Considering the size of the input images, the batch size was set to 12.

\subsection{Evaluation}

To assess the performance and generalization capabilities of the
domain-specific foundation model, we conducted a series of experiments focused
on semantic segmentation tasks. The evaluation metric used was the F1 score,
which balances precision and recall and is well-suited for multi-class
segmentation problems with class imbalance.

Evaluation was performed on the BASE test subset and the full REALITY and DRONE
datasets, which were not used during training. These datasets encompass diverse
acquisition conditions, including variations in camera types, geographic
locations, and capture protocols, enabling a robust assessment of domain
generalization.

All reported metrics correspond to the best-performing model checkpoint,
selected based on the validation performance on the 2019A2 subset.

Moreover, given the importance of accurate species identification
in herbicide trials, we conducted detailed analyses for each dataset. This
included examining confusion matrices and analyzing the relationship between
training data quantity (annotated pixels per class) and model performance.

To further analyze annotation efficiency, we performed an incremental training
experiment where the amount of labeled data was progressively reduced. This
allowed us to evaluate the model's performance in low-annotation regimes and
its potential to reduce the manual labeling burden. Specifically, we selected
an equal number of images from each collection, using increments of 25, 50, 75,
100, and 147 images per collection. This resulted in total training set sizes
of 50, 100, 150, 200, and 294 images, respectively. The maximum value of 147
was chosen to match the size of the smallest collection (2019A1), ensuring that
the training dataset remains balanced across collections (see Table
\ref{tab:real_field_datasets}).

Additionally, to gain insight into the learned representations, we visualized
the first three principal components of the features extracted by the final
block of the transformer encoder. Following the approach of
\cite{oquab_dinov2_2024}, Principal Component Analysis (PCA) was applied to
reduce the feature dimensionality, and each component was mapped to an RGB
channel to generate interpretable visualizations. Background regions were
removed using the provided vegetation segmentation masks.

\section{Results}
\label{sec:results}

In this section, we evaluate the performance of the domain-specific foundation
model across multiple semantic segmentation tasks. We report results on
datasets with varying acquisition conditions, analyze the impact of reducing
labeled data, and visualize the learned representations to better understand
the model's internal features.

\subsection{Evaluation on BASE and REALITY datasets}

A comprehensive evaluation of the model was carried out using two distinct
datasets to capture both standard performance and generalization under domain
shift. The BASE dataset (see Section \ref{subsec:base_data}) served as the
foundation for training and initial testing, following the previously described
split (see Section \ref{subsec:sup_ft}).

To further assess the model's robustness in more realistic and diverse
conditions, we evaluated it using the REALITY dataset, which was collected in
2023 across Germany, Spain, and the U.S. Unlike BASE, this dataset introduces
significant domain shifts. Therefore, by training on BASE and testing on
REALITY, we ensure an unbiased evaluation that reflects the model's ability to
generalize to previously unseen environments and conditions.

Table \ref{tab:results_reality_comparative} provides a comparative analysis of
the F1 scores achieved by the baseline general-purpose foundation models
(DINOv2, DINOv3\_web, and DINOv3\_sat) and the developed domain-specific
variants. While vegetation classification remains consistently high
across all models and settings ($F1 = 1.0$), notable improvements are observed
in
species and damage classification when using domain-specific models.

Specifically, when comparing the best-performing baseline (DINOv2) with the
best domain-specific model (DINOv2\_FT\_224). The F1 score for species
classification increased from 0.91 to 0.94 on the BASE dataset and from 0.56 to
0.66 on the REALITY dataset. For damage classification, the F1 score improved
from 0.26 to 0.33 on BASE and from 0.17 to 0.27 on REALITY.

\begin{table}[ht]
  \centering
  \begin{tabular}{lllll}
    \hline
    Trained/Tested & Encoder/Decoder                  & F1           & F1
                   &
    F1
    \\
                   &                                  & Vegetation   & Species
                   &
    Damage
    \\
    \hline

    BASE/BASE      & DINOv2$/$MultiScaleHead          & \textbf{1.0} &
    0.91           &
    0.26
    \\
    BASE/BASE      & DINOv3\_web$/$MultiScaleHead     & \textbf{1.0} &
    0.91           &
    0.15
    \\
    BASE/BASE      & DINOv3\_sat$/$MultiScaleHead     & \textbf{1.0} &
    0.84           &
    0.20
    \\
    BASE/BASE      & DINOv2\_FT\_518$/$MultiScaleHead & \textbf{1.0} &
    0.92           &
    0.32
    \\
    BASE/BASE      & DINOv2\_FT\_224$/$MultiScaleHead & \textbf{1.0} &
    \textbf{0.94}  &
    \textbf{0.33}
    \\
    \hline
    BASE/REALITY   & DINOv2$/$MultiScaleHead          & \textbf{1.0} &
    0.56           &
    0.17
    \\
    BASE/REALITY   & DINOv3\_web$/$MultiScaleHead     & \textbf{1.0} &
    0.55           &
    0.11
    \\
    BASE/REALITY   & DINOv3\_sat$/$MultiScaleHead     & \textbf{1.0} &
    0.33           &
    0.11
    \\
    BASE/REALITY   & DINOv2\_FT\_518$/$MultiScaleHead & \textbf{1.0} &
    0.64           &
    0.25
    \\
    BASE/REALITY   & DINOv2\_FT\_224$/$MultiScaleHead & \textbf{1.0} &
    \textbf{0.66}  &
    \textbf{0.27}
    \\
    \hline
  \end{tabular}
  \caption{Metrics of model architectures trained with the BASE dataset (2019)
    and evaluated against the testing subset of the BASE dataset (2019A2) or
    the
    REALITY (2023) dataset.}
  \label{tab:results_reality_comparative}
\end{table}

\begin{figure*}[ht]
  \centering
  \includegraphics[width=15cm]{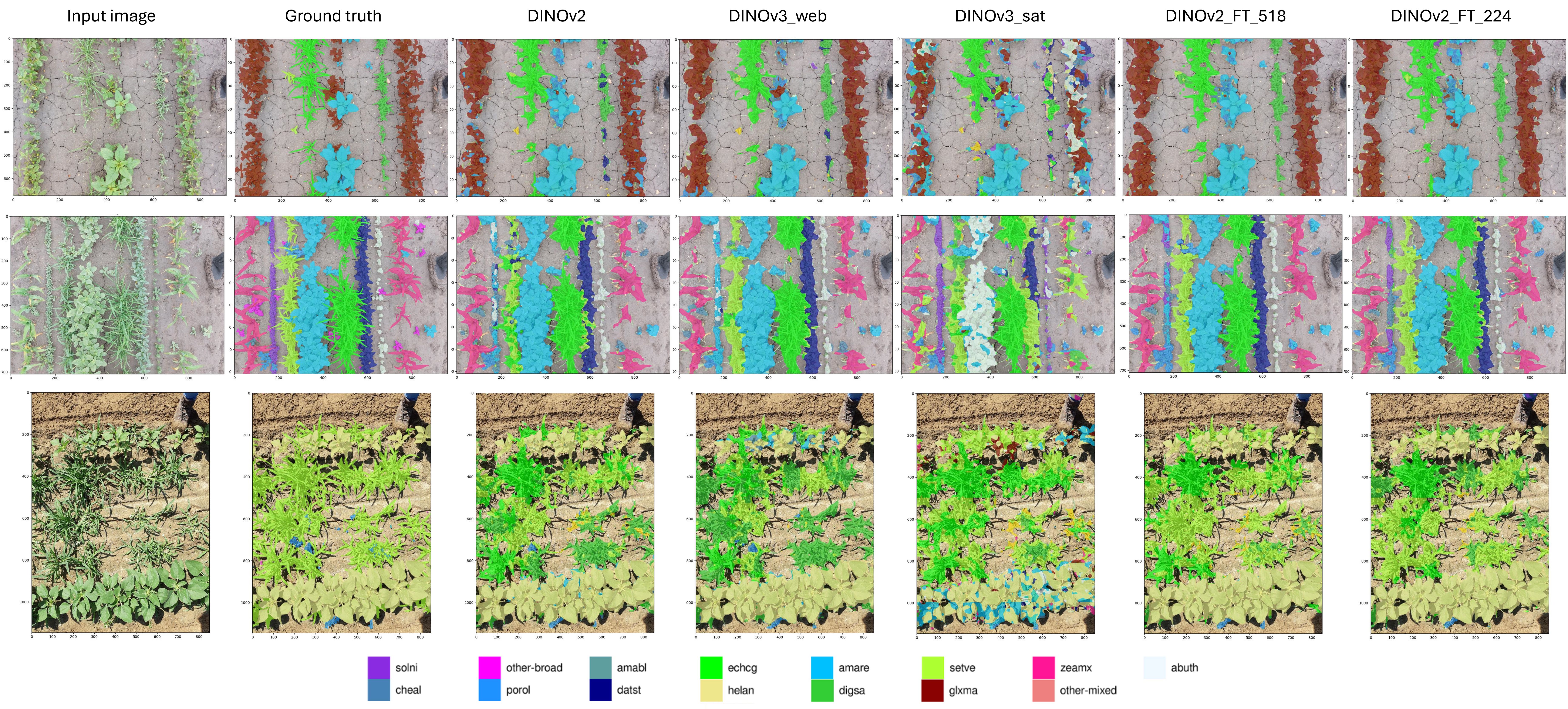}
  \caption{Species predictions generated by each evaluated encoder across
    different crop fields. Model trained on the BASE dataset and tested on the
    REALITY dataset: top) \textit{Glycine max} field, middle) \textit{Zea mays}
    field, bottom) \textit{Helianthus annuus} field.}
  \label{fig:segmentation_results_reality}
\end{figure*}

Figure \ref{fig:segmentation_results_reality} provides qualitative examples of
species identification in the REALITY dataset across different model backbones.
While all models are capable of accurately detecting vegetation regions,
species-level classification remains more challenging. The DINOv2 models
consistently yield the best results, with the fine-tuned variants
(DINOv2\_FT\_518 and DINOv2\_FT\_224) showing slightly improved
performance, particularly in identifying less dominant weed species.

\subsubsection{Species identification}
\begin{figure}[ht]
  \centering
  \includegraphics[width=8cm]{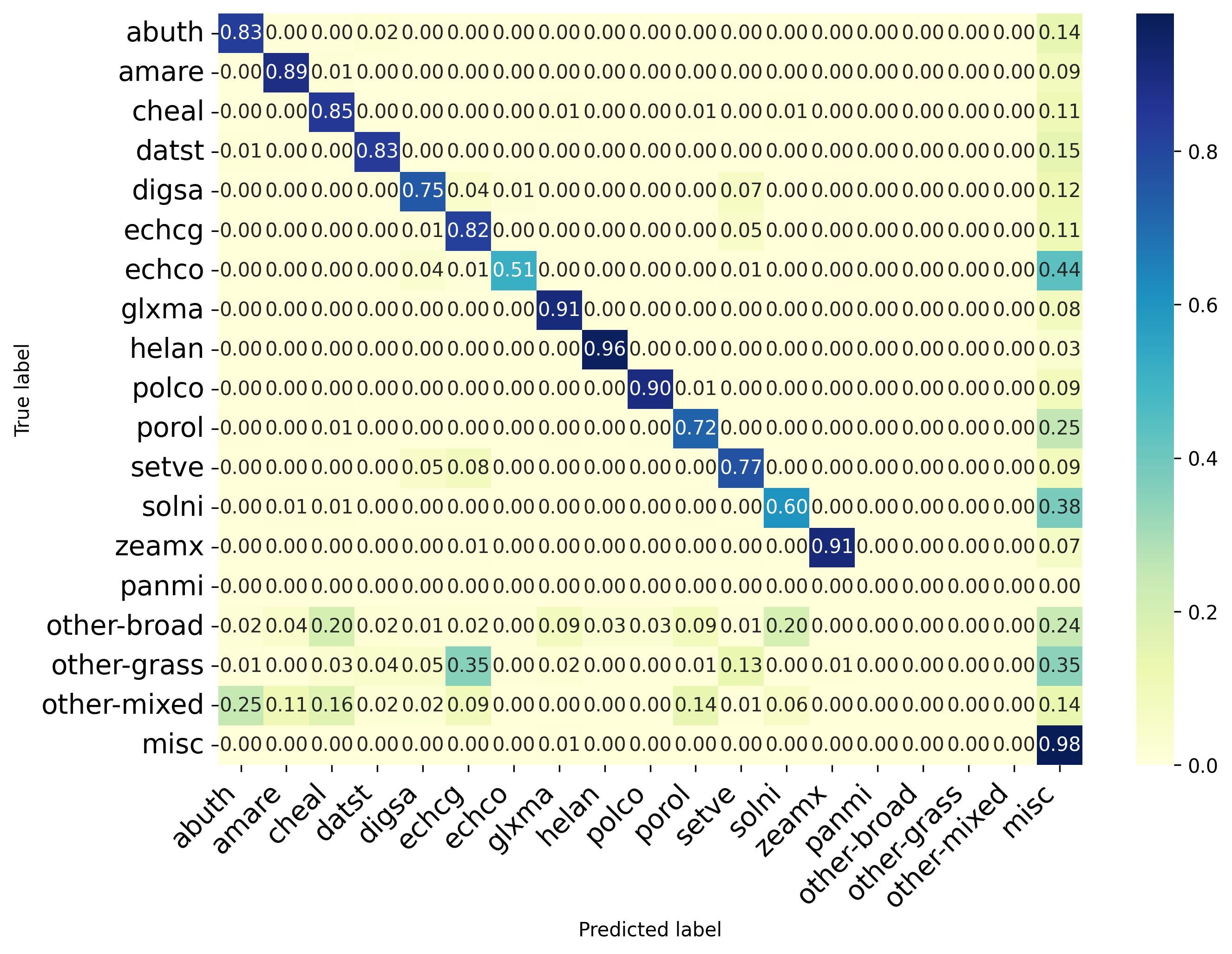}
  \includegraphics[width=8cm]{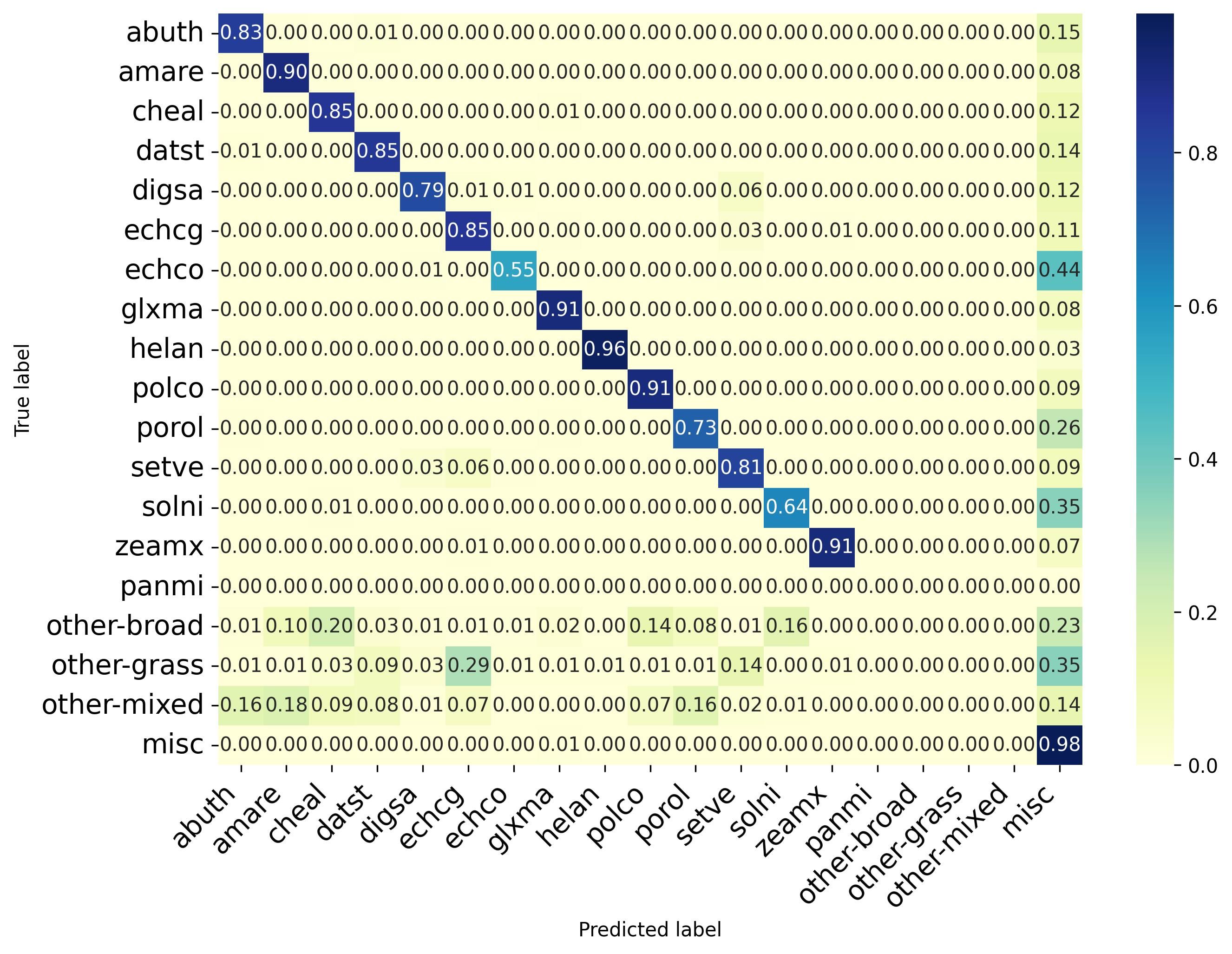}
  \caption{Comparative of the confusion matrices for species identification.
    Trained and tested on the BASE dataset: left) DINOv2 (Best baseline) right)
    DINOv2\_FT\_224 (Best domain-specific foundation model).}
  \label{fig:species_cm_test}
\end{figure}

Figure \ref{fig:species_cm_test} shows the difference between the best
performing baseline (DINOv2) and the best domain-specific model
(DINOv2\_FT\_224) tested in the BASE dataset. Overall, the number of
misclassifications is reduced in the fine-tuned model, with most predictions
remaining unchanged or showing slight improvements. The confusion matrix
suggests that DINOv2\_FT\_224 achieves more consistent accuracy across species,
particularly by correcting cases that were previously misclassified by the
baseline.

\begin{figure}[ht]
  \centering
  \includegraphics[width=8cm]{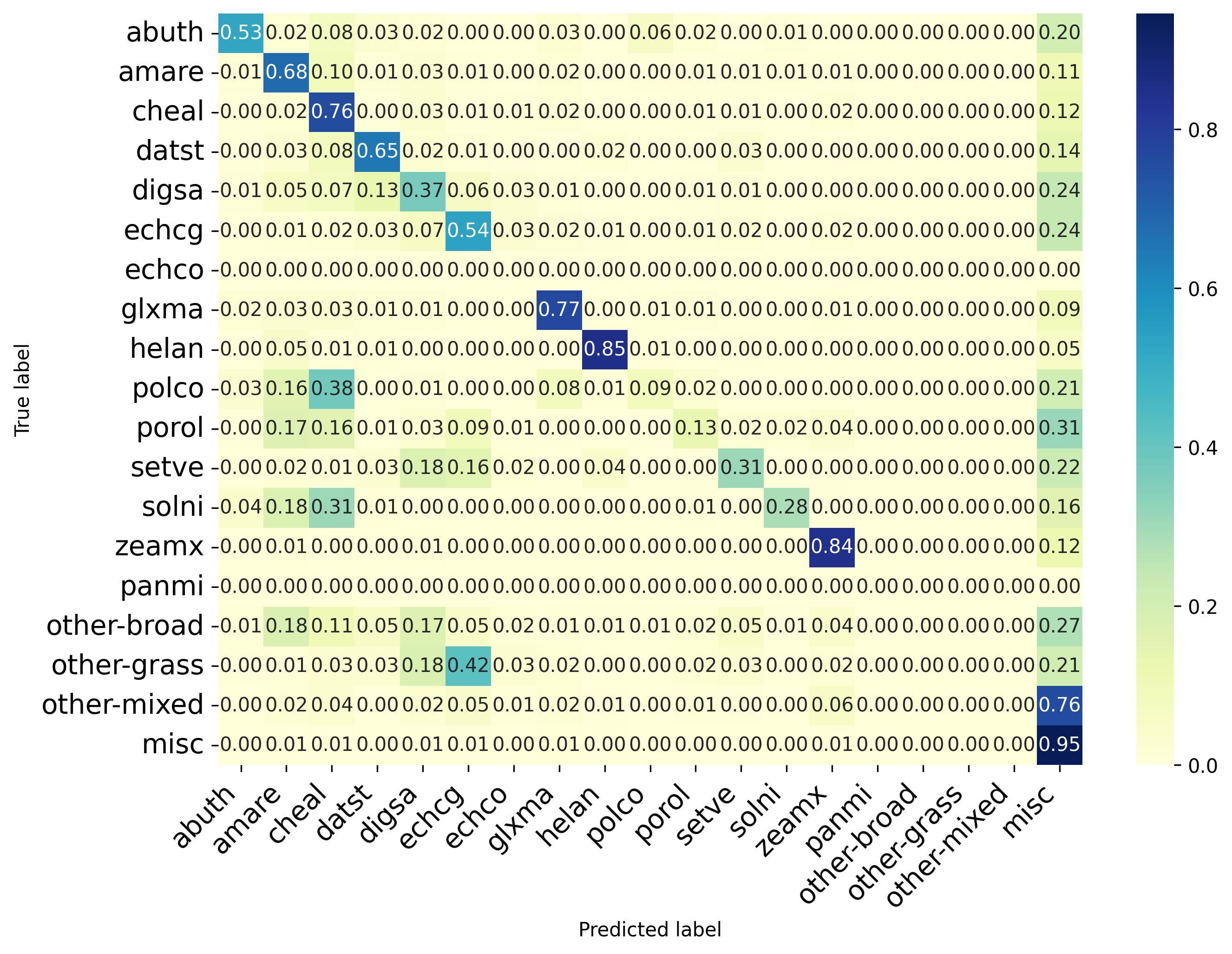}
  \includegraphics[width=8cm]{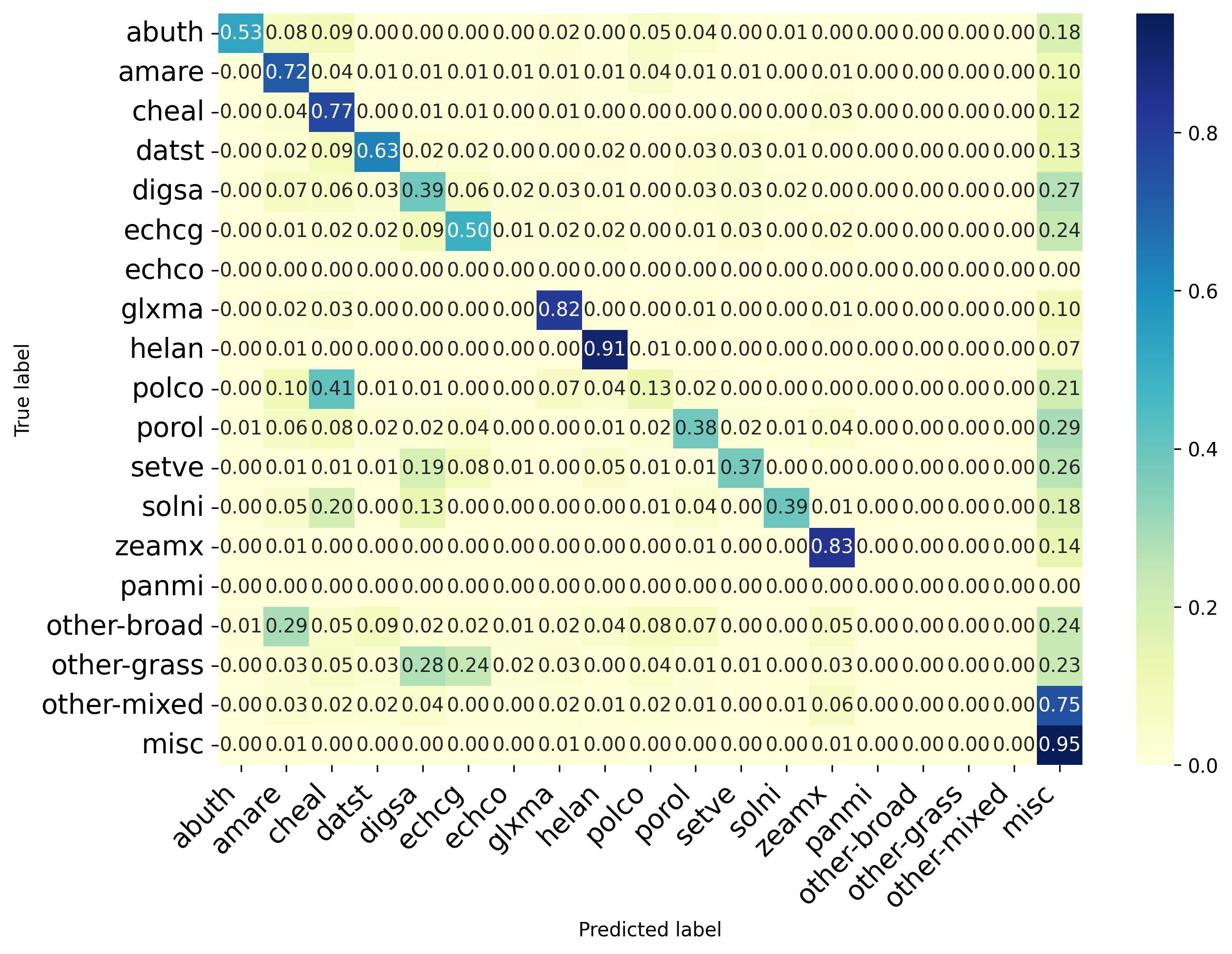}
  \caption{Comparative of the confusion matrices for species identification.
    Trained on the BASE dataset and tested on the REALITY dataset: left) DINOv2
    (Best baseline) right) DINOv2\_FT\_224 (Best domain-specific foundation
    model).}
  \label{fig:species_cm_reality}
\end{figure}

To complement the evaluation on the BASE dataset, Figure
\ref{fig:species_cm_reality} presents the confusion matrices obtained from
testing in the REALITY dataset. As observed previously, the domain-specific
model (DINOv2\_FT\_224) continues to outperform the baseline (DINOv2),
showing stronger diagonal dominance and fewer misclassifications across most
categories. Notable improvements are seen in classes such as \textit{porol}
(from 0.13 to 0.38) and \textit{solni} (from 0.28 to 0.39), while persistent
confusion in \textit{polco} (from 0.09 to 0.13) suggests that certain species
remain challenging regardless of training domain.

\begin{figure}[ht]
  \centering
  \includegraphics[width=8cm]{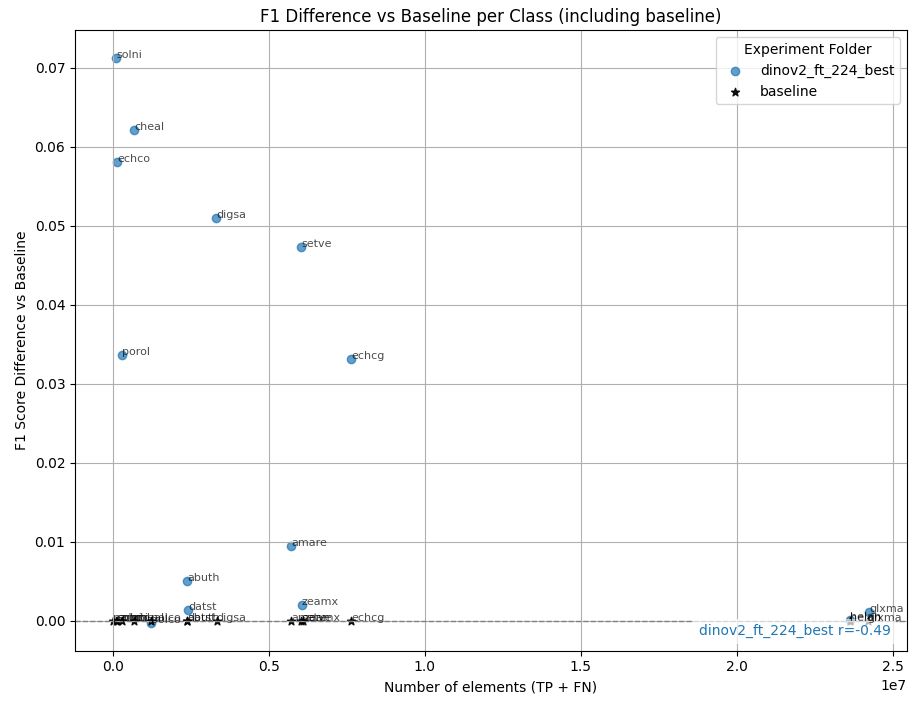}
  \includegraphics[width=8cm]{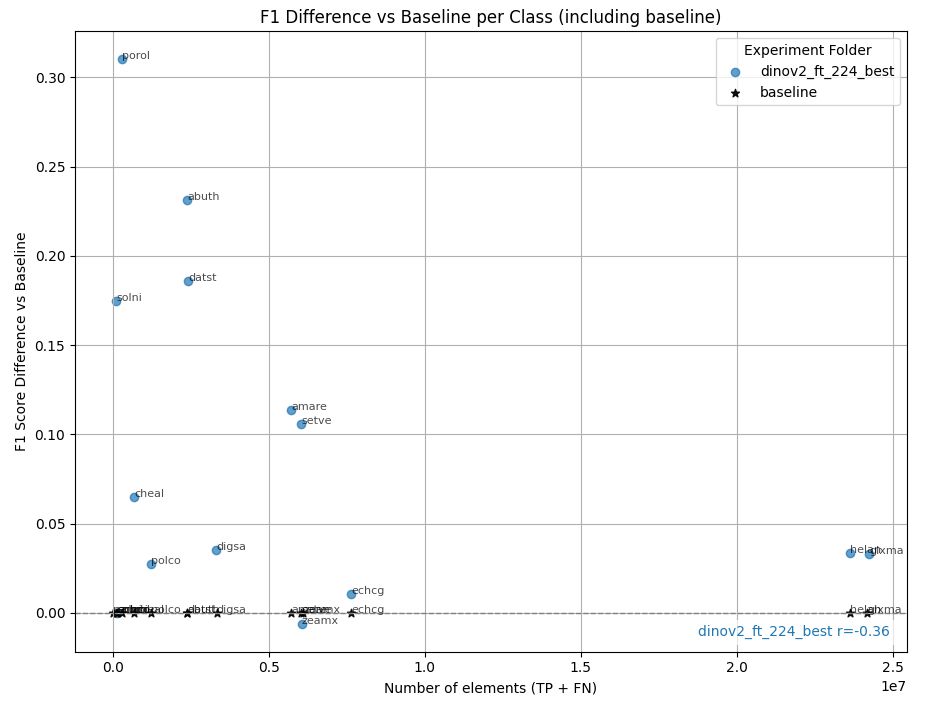}
  \caption{Class-wise comparison of F1 score improvements over the baseline
    (DINOv2) for species identification. The plot illustrates how model
    performance
    varies depending on the number of annotated pixels per class used during
    training. Trained on the BASE dataset and tested on the : left) BASE
    dataset, and right) REALITY dataset.}
  \label{fig:pixel_annot_difference_species}
\end{figure}

Furthermore, to understand the effect of the quantity of annotations, a
class-wise
comparison of F1 score improvements over the best performing general-purpose
model (DINOv2) and the best domain-specific model (DINOv2\_FT\_224) has been
performed. As illustrated in Figure \ref{fig:pixel_annot_difference_species},
the most pronounced improvements are observed in classes with fewer annotated
pixels. In contrast, classes with abundant annotations tend to show minimal
change, either slight improvements or minimal decreases. This trend is
consistent across both the BASE and REALITY evaluations, with improvements
being more pronounced in the REALITY dataset. Specifically, \textit{solni}
shows the greatest improvement in the BASE evaluation, while \textit{porol}
stands out in REALITY.

\subsection{Evaluation on the DRONE Dataset: Assessing Domain Shift Robustness}

To evaluate how well the model performs under significant domain shifts, we
tested it using the DRONE dataset. This dataset differs significantly from the
training data in several aspects: the type of image source (drone sensor
instead of digital cameras), the time of data collection (2024 vs. 2019), and
the surrounding environmental conditions. These differences create a
challenging and unbiased scenario to assess the model's ability to generalize.
Since manual damage annotations are not available for this dataset, our
analysis focuses solely on species identification and vegetation
classification.

Table~\ref{tab:results_reality_comparative_drone} presents a comparative
analysis of the F1 scores obtained by the best-performing baseline (DINOv2) and
the domain-specific variants on the drones dataset. Despite the challenging
conditions, all models maintain perfect F1 scores ($F1 = 1.0$) for vegetation
classification.

In contrast, species classification shows clear sensitivity to domain shift.
DINOv2 achieves an F1 score of 0.49, while domain-specific fine-tuning leads to
consistent improvements, with the best-performing model (DINOv2\_FT\_224)
reaching 0.60.
\begin{table}[ht]
  \centering
  \begin{tabular}{llll}
    \hline
    Trained/Tested & Encoder/Decoder                  & F1           & F1
    \\
                   &                                  & Vegetation   & Species
    \\
    \hline
    BASE/DRONE     & DINOv2$/$MultiScaleHead          & \textbf{1.0} &
    0.49
    \\
    BASE/DRONE     & DINOv2\_FT\_518$/$MultiScaleHead & \textbf{1.0} &
    0.54
    \\
    BASE/DRONE     & DINOv2\_FT\_224$/$MultiScaleHead & \textbf{1.0} &
    \textbf{0.60}
    \\
    \hline
  \end{tabular}
  \caption{Performance of models trained on the BASE dataset (2019) and
    evaluated on the DRONE (2024) dataset.}
  \label{tab:results_reality_comparative_drone}
\end{table}

\begin{figure*}[ht]
  \centering
  \includegraphics[width=15cm]{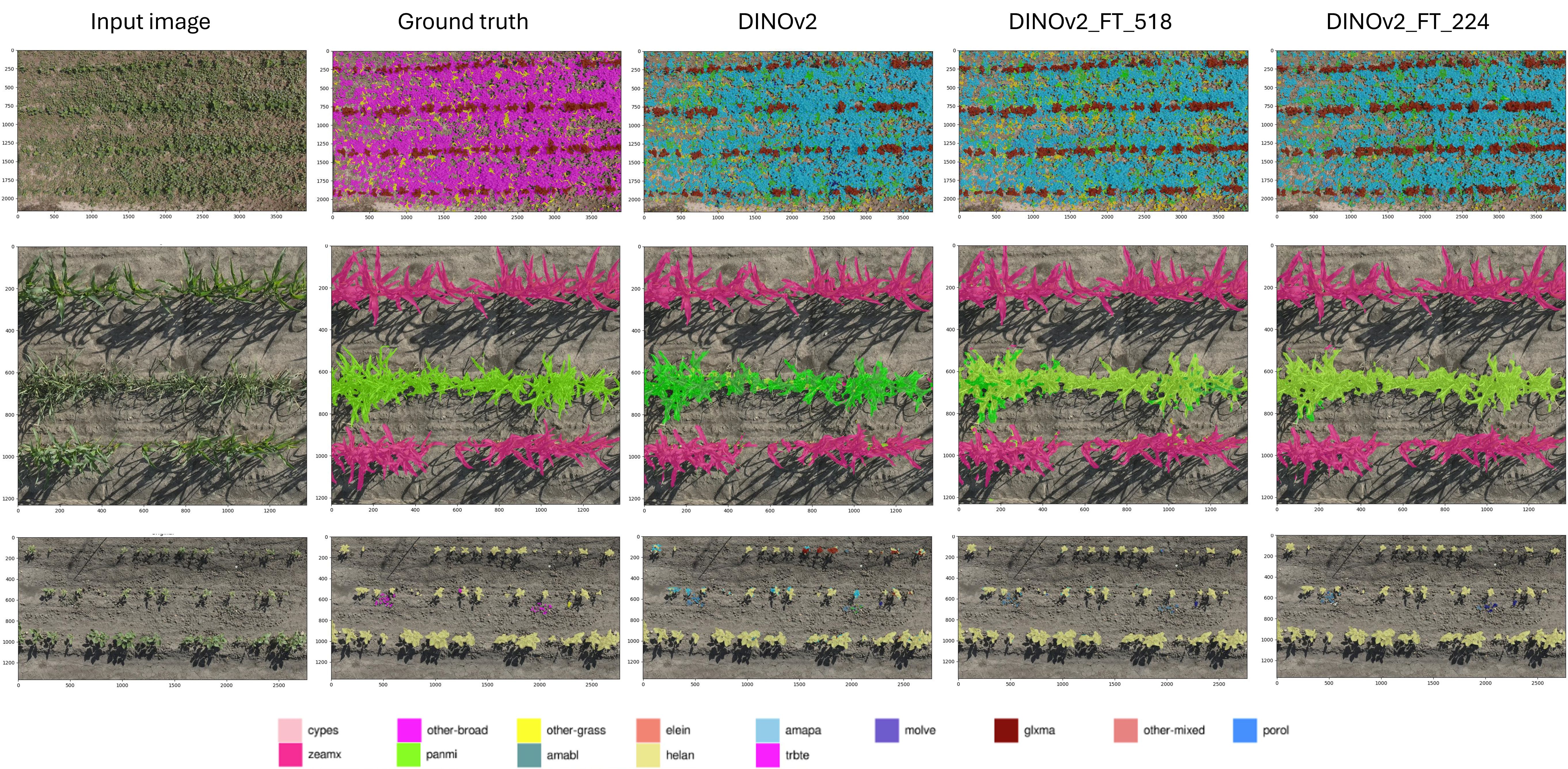}
  \caption{Species predictions generated by each evaluated encoder across
    different crop fields. Model trained on the BASE dataset and tested on the
    DRONE dataset: top) \textit{Glycine max} field, middle) \textit{Zea mays}
    field, bottom) \textit{Helianthus annuus} field.}
  \label{fig:segmentation_results_drone}
\end{figure*}

Examples of species identification in the DRONE dataset are shown in Figure
\ref{fig:segmentation_results_drone}. Although all models can accurately detect
vegetation areas, distinguishing species is still more difficult. The most
notable errors are associated with species labeled as "other", which typically
occur when species are either unfamiliar to technicians or not included among
the model's supported categories (such as other-grass, other-broad, and
other-mixed).

\subsubsection{Species identification}
\begin{figure}[ht]
  \centering
  \includegraphics[width=8cm]{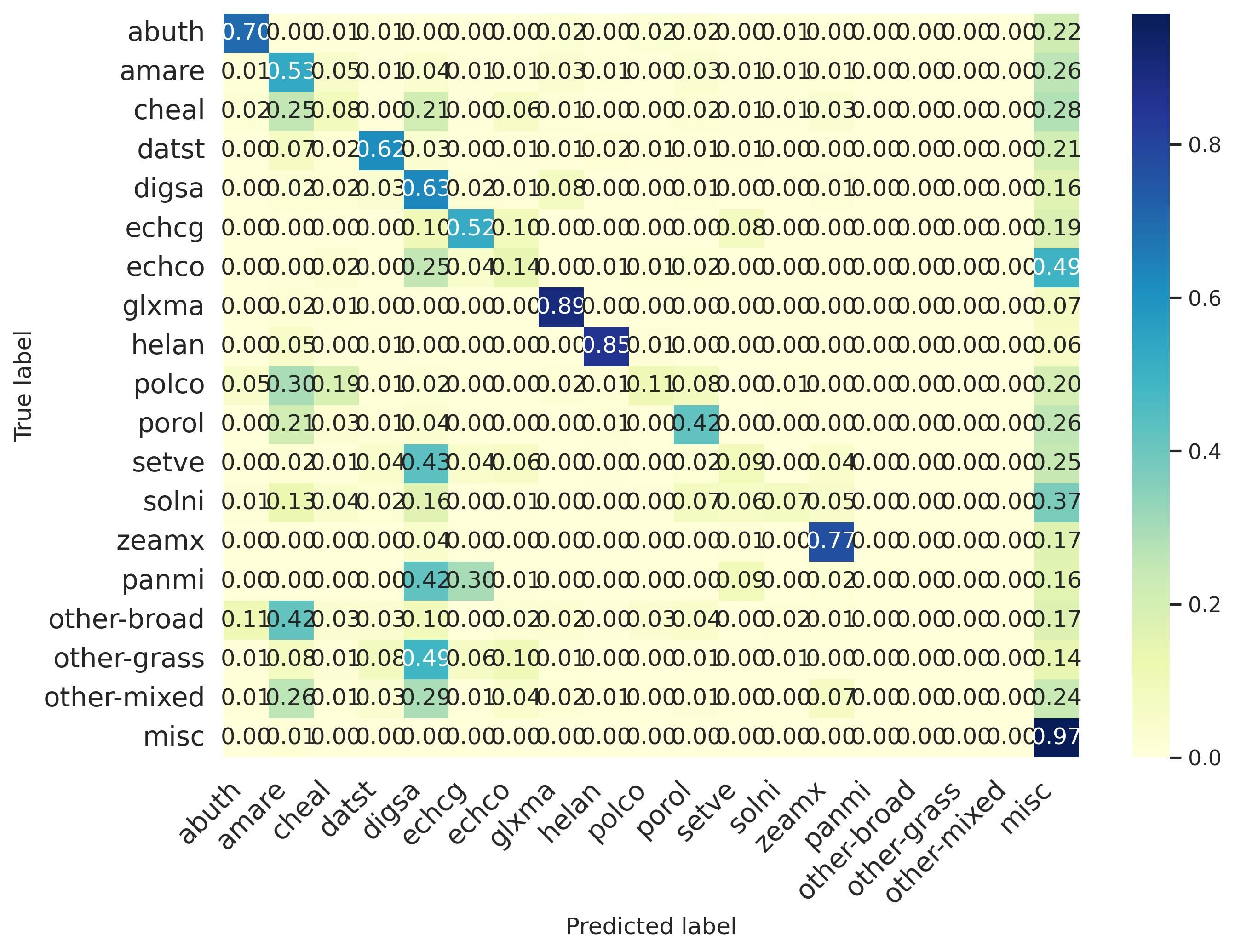}
  \includegraphics[width=8cm]{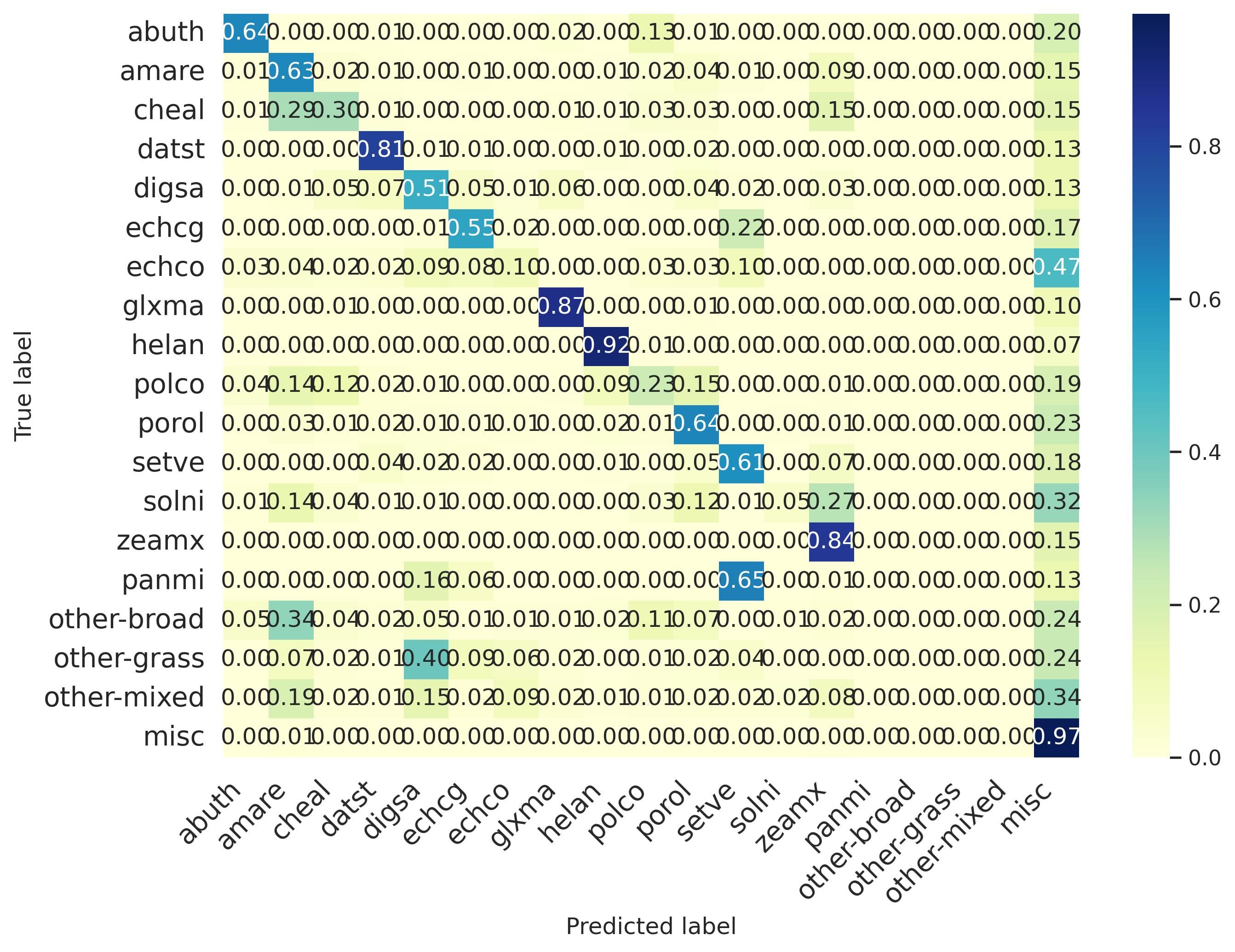}
  \caption{Comparative of the confusion matrices for species identification.
    Trained on the BASE dataset and tested on the DRONE dataset: left) DINOv2
    (Best baseline) right) DINOv2\_FT\_224 (Best domain-specific foundation
    model).}
  \label{fig:species_cm_drone}
\end{figure}

Figure \ref{fig:species_cm_drone} compares the top-performing general-purpose
model (DINOv2) with the best domain-adapted model (DINOv2\_FT\_224) on the
DRONE dataset. As seen in previous evaluations, the domain-specific model
demonstrates a notable reduction in misclassifications
across most species. Improvements are particularly evident in categories such
as \textit{cheal} and \textit{setve}, while some confusion persists for classes
like \textit{solni}, indicating ongoing challenges for certain species.

\begin{figure}[ht]
  \centering
  \includegraphics[width=9cm]{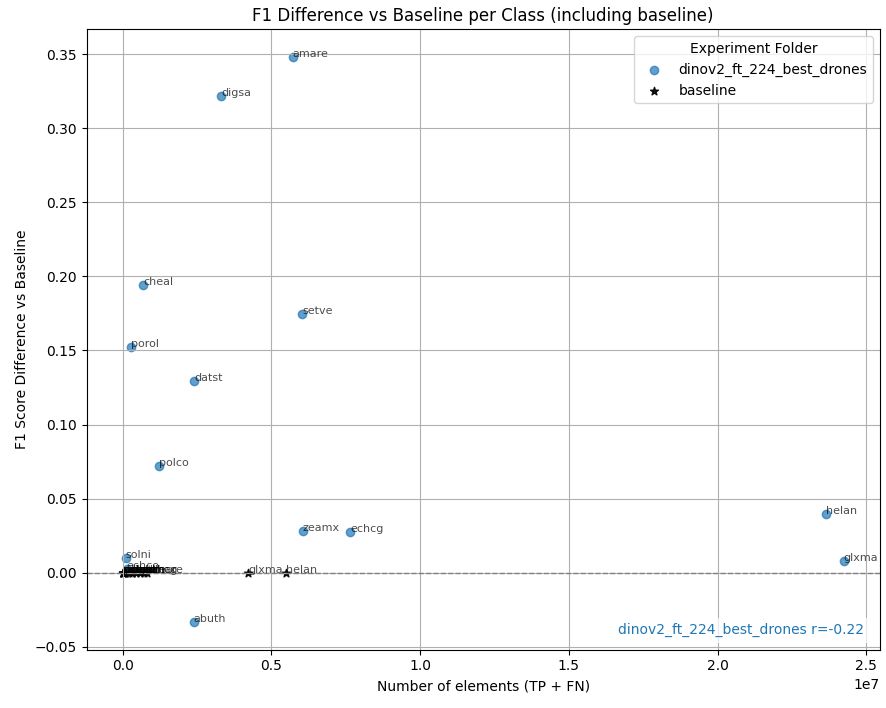}
  \caption{Class-wise comparison of F1 score improvements over the baseline
    (DINOv2) for species identification. The plot illustrates how model
    performance
    varies depending on the number of annotated pixels per class used during
    training. Trained on the BASE dataset and tested on the DRONE dataset.}
  \label{fig:pixel_annot_difference_species_drones}
\end{figure}

In addition, Figure \ref{fig:pixel_annot_difference_species_drones} explores
how the amount of annotated data per class affects model performance on the
DRONE dataset. This figure compares the class-wise F1 score improvements
achieved by the domain-adapted model (DINOv2\_FT\_224) over the baseline
(DINOv2). The results show that the largest gains are concentrated in species
with fewer annotated pixels. In contrast, classes with more extensive
annotations tend to exhibit only minor changes, with either slight improvements
or negligible declines.

\subsection{Evaluation with Limited Labeled Data}
As seen in the previous sections, the domain-specific foundation models show
substantial gains in low-data regimes (see Figures
\ref{fig:pixel_annot_difference_species} and
\ref{fig:pixel_annot_difference_species_drones}). Therefore, to furhter explore
their potential, we evaluate the performance under extremely limited labeled
data.

\begin{figure}[ht]
  \centering
  \includegraphics[width=8cm]{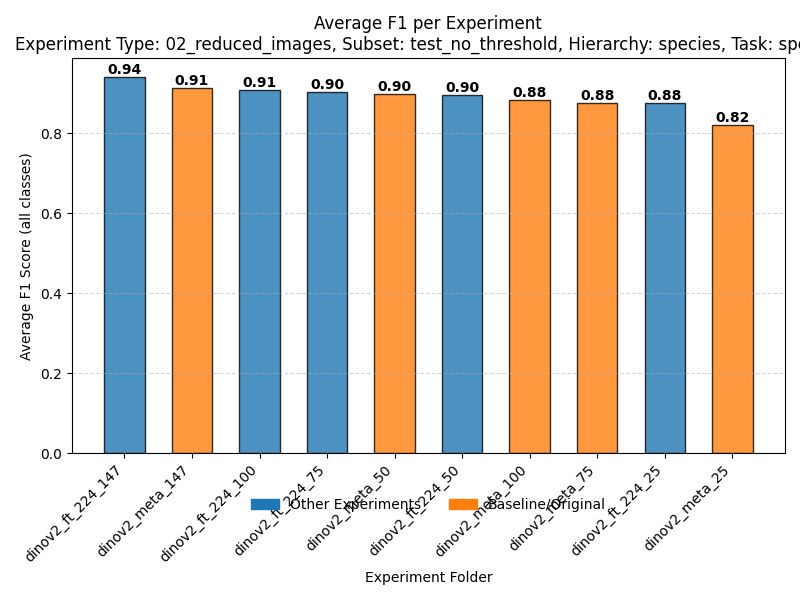}
  \includegraphics[width=8cm]{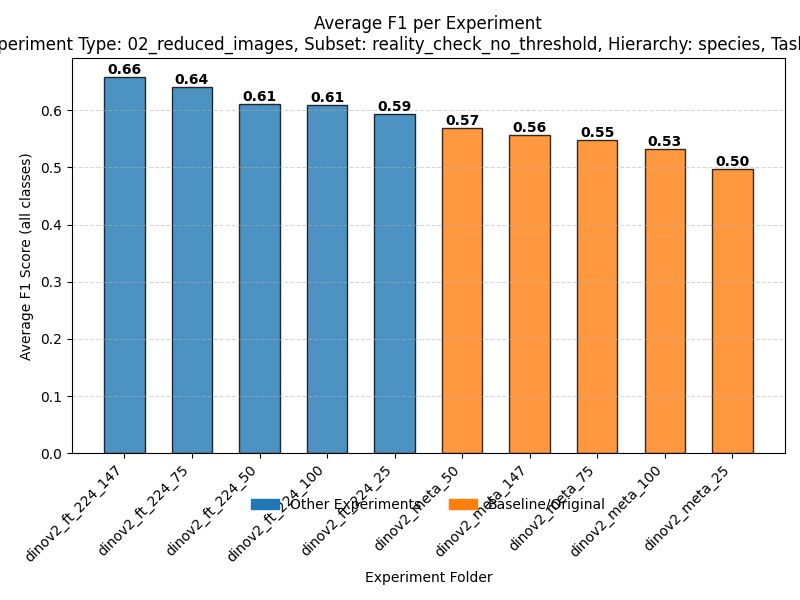}
  \caption{Comparison of F1 score in species identification across models
    trained with varying amounts
    of labeled data per collection. The plot illustrates how performance varies
    depending on the volume of annotated training data. Trained on the BASE
    dataset and tested on the : left) BASE dataset, and right) REALITY
    dataset.}
  \label{fig:reduced_num_labels}
\end{figure}

Figure \ref{fig:reduced_num_labels} compares the best-performing baseline
(DINOv2) and the top domain-specific model (DINOv2\_FT\_224), each trained on
progressively smaller subsets of the BASE dataset. Evaluation is conducted on
both the BASE and REALITY datasets to assess generalization accross domains.

In the BASE evaluation (left), both models mantain relatively high performance
despite the reduced training dataset, with DINOv2\_FT\_224 consistently
outperforming or matching its counterpat accross all configurations.

In the REALITY evaluation (right), the advantage of domain-specific pretraining
becomes especially pronounced. At every training size, DINOv2\_FT\_224
consistently outperforms DINOv2, maintaining a clear performance advantage
regardless of the amount of labeled data. Notably, with only 25 images
per collection (50 total), the domain-specific model achieves an F1 score of
0.59, outperforming both its baseline variant ($F1=0.5$) and even the
one trained with 147 images per collection (294 total), which only reaches
0.56.

\subsection{Visual analysis}

\begin{figure*}[ht]
  \centering
  \includegraphics[width=15cm]{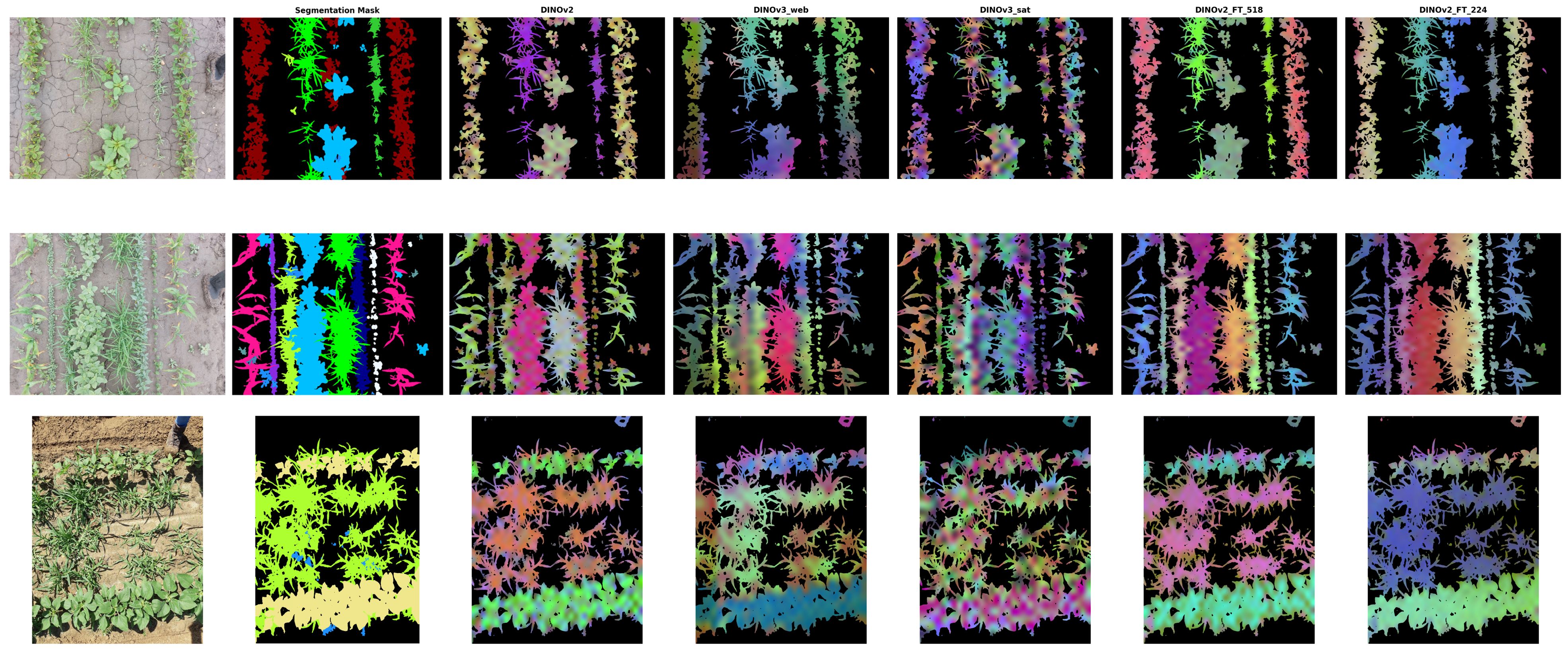}
  \caption{Visualization of the features extracted by each of the evaluated
    encoders across different crop fields. To generate the colored
    visualization, PCA is performed on each of the samples, and the first three
    components are mapped to the RGB channels. The background is removed using
    the provided vegetation segmentation mask. Features are shown for: top)
    \textit{Glycine max} field, middle) \textit{Zea mays} field, bottom)
    \textit{Helianthus annuus} field.}
  \label{fig:pca_visualization}
\end{figure*}

Herbicide trial images differ notably from natural ones in their visual
properties. They tend to have a more uniform color distribution, and objects in
the foreground may not be easily distinguishable due to overlapping elements or
background with similar appearances.  To separate the background, vegetation
segmentation masks provided with the dataset are utilized. For each image whose
features are to be visualized, we perform Principal Component Analysis (PCA),
and map the first three principal components to the RGB channels to
generate the colored visualizations. The same three samples shown in Figure
\ref{fig:segmentation_results_reality} are selected, and their corresponding
feature visualizations, extracted by each encoder, are presented in Figure
\ref{fig:pca_visualization}.

The visualizations in Figure \ref{fig:pca_visualization} show that all models
effectively capture vegetation structure. DINOv2-based encoders are
particularly effective at distinguishing between different species of crops and
weeds. The fine-tuned DINOv2 models exhibit improved intra-image consistency,
assigning similar colors to plants of the same species within a single image,
unlike the non-fine-tuned version, which produces less stable species-level
representations.


\section{Discussion}
\label{sec:Discussion}

This study introduces a domain-specific vision foundation model tailored for
herbicide trials, trained using a self-supervised learning approach on a large,
curated agricultural dataset. The results demonstrate that domain-specific
pretraining yields rich and transferable representations, consistently
outperforming general-purpose foundation models across semantic segmentation
tasks involving multiple plant species and damage types.

On the BASE dataset, which represents low domain shift conditions, the
domain-specific model achieves notable improvements in F1 score for both
species identification (from 0.91 to 0.94) and damage classification (from 0.26
to 0.33).

These gains are even more pronounced under domain shift scenarios. On the
REALITY dataset, which introduces temporal and environmental variability, the
model maintains robust performance, with species identification improving from
0.56 to 0.66 and damage classification from 0.17 to 0.27. These results
highlight the importance of domain adaptation. When examining the
impact of the backbone on generalization, it becomes evident that
general-purpose
models suffer more pronounced performance degradation under domain shift,
whereas domain-specific models exhibit greater resilience and adaptability to
new scenarios.

Moreover, under more extreme domain shifts, including changes in camera types,
geographic locations and capture protocols, the model still demonstrates strong
generalization, improving species identification F1 score from 0.49 to 0.60.
These results further confirm the practical value of domain-specific
pretraining. The shift in sensor modality and environmental context introduced
by drone data reveals the limitations of general-purpose models and the
enhanced
adaptability of their domain-specific counterparts.

Given the high cost of manual annotation in agricultural settings, we further
examined how the model performs under limited supervision. By
progressively reducing the amount of labeled data, we show that the
domain-specific model achieves high segmentation accuracy even when data is
scarce (see Figure \ref{fig:reduced_num_labels}). On the REALITY dataset, it
achieves a 5.4\% higher F1 score than the general-purpose model while requiring
80\%
fewer labeled samples. These results highlight the annotation efficiency and
generalization capacity of domain-specific models, particularly in unseen
scenarios. This is particularly relevant in agricultural contexts, where manual
annotation is both labor-intensive and costly. The ability to maintain strong
performance with limited supervision offers a scalable and cost-effective
solution for herbicide trial analysis.

An important methodological consideration in this study was how to optimize
pretraining under limited hardware resources. To this end, we trained two
versions of the domain-specific model at different input resolutions. This
approach enabled a larger global batch size, which is beneficial for
self-supervised pretraining (\cite{pmlr-v119-chen20j}). Interestingly, the
model pretrained at lower resolution (224x224) consistently outperformed the
one trained at higher resolution (518x518), despite both surpassing the
baseline (see Tables \ref{tab:results_reality_comparative} and
\ref{tab:results_reality_comparative_drone}). These results suggest that,
within the context of self-supervised pretraining for weed herbicide trials,
increasing batch size may have a greater impact on representation quality than
input resolution alone.

Despite these optimizations, the overall computational demands of pretraining
large vision models remain substantial. Extended training times and the need
for high-end hardware can limit the accessibility and scalability of this
approach. Future work should explore more resource-efficient training
strategies such as Low-Rank Adaptation (LoRA) (\cite{hu_lora_2021}). This
approach could be applied to the supervised fine-tuning, where it has already
demonstrated strong performance in other downstream tasks
(\cite{espejo-garcia_foundation_2025}), or even integrated into
the self-supervised pretraining process to enable more accessible and scalable
domain adaptation. Additionally, while this study focused on species
identification and damage assessment, the proposed approach could be
extended to other downstream agricultural tasks, such as disease detection,
growth stage estimation, or yield prediction.


\section{Conclusions}
\label{sec:conclusions}
This research highlights the advantages of domain-specific vision foundation
models in agricultural segmentation tasks. While performance differences with
general-purpose models are modest under familiar conditions, the benefits
become
more pronounced in the presence of domain shifts and limited annotation.
Through
extensive evaluation, we show that domain-specific models consistently
outperform
their general-purpose counterparts in scenarios that reflect real-world
agricultural variability, maintaining strong performance even with reduced
supervision.

These findings suggest that domain-specific foundation models are better suited
for real-world agricultural scenarios, where variability and limited
supervision are common. Additionally, the model's ability to generalize across
datasets and maintain strong performance with fewer annotations supports its
potential for scalable deployment.

To fully assess the potential of the domain-specific foundation models, more
efficient fine-tuning techniques, such as LoRA, should be explored, to reduce
computational costs and improve adaptability. Moreover, expanding the
evaluation to a broader range of downstream tasks and datasets would help
validate the model's generalization capabilities and further confirm
the advantages of domain-specific foundation models across diverse agricultural
scenarios.

\section*{Author Contributions}
\textbf{LB-D-V}: conceptualization, investigation, software, formal analysis,
methodology, and writing -original draft, review and editing. \textbf{AP}:
conceptualization, investigation, software, formal analysis, methodology,
supervision, and writing -review and editing. \textbf{DM}: conceptualization,
investigation, data curation, software, formal analysis, methodology, and
writing -review and editing. \textbf{MR}: conceptualization, formal analysis,
investigation, methodology, and writing -review and editing. \textbf{EP}:
conceptualization, investigation, methodology, supervision, and writing -review
and editing \textbf{JR}: validation, conceptualization, investigation, and
methodology. \textbf{CJJ}: validation, methodology, investigation, and
conceptualization. \textbf{RN-M}: investigation, methodology, and writing
-review and editing.

\section*{Acknowledgments}

All experiments for this work were conducted using Tecnalia's KATEA computing
cluster.

\section*{Funding}

Some authors have received partial support by the Elkartek Programme, Basque
Government (Spain) (DBASKIN (KK-2025/00012))


\section*{Declaration of competing interest}

LB-D-V, AP and DM were employed by TECNALIA. MR, JR, CJJ, and RN-M were
employed by
BASF SE.

The remaining authors declare that the research was conducted in the absence of
any commercial or financial relationships that could be construed as a
potential conflict of interest.

\section*{Data availability}

The data analyzed in this study is subject to the following
licenses/restrictions. The dataset used in this article has been generated by
the BASF R\&D field research community. It could be made available on
reasonable request for non-commercial research purposes and under an agreement
with BASF. Requests to access these datasets should be directed to
ramon.navarra-mestre@basf.com.


\section*{Declaration of Generative AI and AI-assisted technologies in the
  writing process}

During the preparation of this work the authors used ChatGPT and Copilot in
order to improve language and readability. After using these tools, the authors
reviewed and edited the content as needed and take full responsibility for the
content of the publication.

\FloatBarrier

\bibliography{mybibfile.bib}


\newpage

\end{document}